# Distributed Application of Guideline-Based Decision Support through Mobile Devices: Implementation and Evaluation

Erez Shalom[1], Ayelet Goldstein[1], Elior Ariel[1], Moshe Sheinberger[1], Valerie Jones[2], Boris Van Schooten[3], and Yuval Shahar[1]

[1]The Medical Informatics Research Center,
Department of Software and Information System Engineering,
Ben Gurion University of the Negev, Israel
{erezsh, yshahar, gayelet, eliorar, sheinmos}@bgu.ac.il

[2]University of Twente, The Netherlands
V.M.Jones@utwente.nl

[3]Roessingh Research and Development, Enschede, The Netherlands
b.vanschooten@rrd.nl

**Corresponding Author:**

Erez Shalom
Medical Informatics Research Center
Department of Software and Information Systems Engineering
Ben-Gurion University of the Negev
P.O.B. 653
Beer-Sheva 8410501, Israel
e-mail: erezsh@bgu.ac.il

Telephone: +972-54-431-2565

# Abstract

**Background**

Traditionally *guideline* (GL)-based *Decision Support Systems* (DSSs) use a centralized infrastructure to generate recommendations to care providers, rather than to patients at home. However, managing patients at home is often preferable, reducing costs and empowering patients. Thus, we wanted to explore an option in which patients, in particular chronic patients, might be assisted by a local DSS, which interacts as needed with the central DSS engine, to manage their disease outside the standard clinical settings.

**Objectives**

To design, implement, and demonstrate the technical and clinical feasibility of a new architecture for a distributed DSS that provides patients with evidence-based guidance, offered through applications running on the patients' mobile devices, monitoring and reacting to changes in the patient's personal environment, and providing the patients with appropriate GL-based alerts and personalized recommendations; and increase the overall robustness of the distributed application of the GL.

**Methods**

We have designed and implemented a novel *projection–callback* (PCB) model, in which small portions of the evidence-based guideline's procedural knowledge are *projected* from a projection engine within the central DSS server, to a local DSS that resides on each patient's mobile device. The local DSS then applies that knowledge using the mobile device's local resources. Furthermore, the projection engine generates guideline projections that are adapted to the patient's previously defined preferences and, implicitly, to the patient's current context, in a manner that is embodied in the projected therapy plans. When appropriate, as defined by a temporal pattern within the projected plan, the local DSS *calls back* the central DSS, requesting further assistance, possibly another projection. To support the new model, the initial specification of the GL includes two levels: one for the central DSS, and one for the local DSS. We have implemented a distributed GL-based DSS using the projection–callback model within the *MobiGuide* EU project, which automatically manages chronic patients at home using sensors on the patients and their mobile phone. We assessed the new GL specification process, by specifying two complex guidelines: for Gestational Diabetes Mellitus, and for Atrial Fibrillation. We evaluated the new computational architecture by applying the two GLs to patients in Spain and Italy, respectively.

**Results**

The specification using the new projection-callback model was found to be quite feasible. We found significant differences between the distributed versions of the two GLs, suggesting further research directions and possibly additional ways to analyze and characterize guidelines. Applying the two GLs to the two patient populations proved highly feasible as well. The mean time between the central and local interactions was quite different for the two GLs: 3.95±1.95 days in the case of the gestational diabetes domain, and 23.80±12.47 days, in the case of the atrial fibrillation domain, probably corresponding to the difference in the distributed specifications of the two GLs. Most of the interaction types were due to projections to the mDSS (83%); others were data notifications, mostly to change context (17%). Some of the data notifications were triggered due to technical errors. The robustness of the distributed architecture was demonstrated through the successful recovery from multiple crashes of the local DSS.

**Conclusions**

The new projection-callback model has been demonstrated to be feasible, from specification to distributed application. Different GLs might significantly differ, however, in their distributed specification and application characteristics.

Distributed medical DSSs can facilitate the remote management of chronic patients by enabling the central DSSs to delegate, in a dynamic fashion, determined by the patient's context, much of the monitoring and treatment management decisions to the mobile device. Such a local management might offer multiple advantages, in particular, by keeping the patients in their home environment, while still maintaining, through the projection-callback mechanism, several of the advantages of a central DSS, such as access to the patient's longitudinal record, and to an up-to-date evidence-based GL repository.

**Keywords**

Distributed Computing, Clinical Guidelines, Clinical Decision Support System, Knowledge engineering.

# 1 Introduction

*Clinical Guidelines* (GLs) are a well-established method for enhancing the quality of care and for reducing costs through the use of evidence-based medical care [1]. Over the past two decades, several automated GL-based application frameworks were developed to provide clinical decision support to care providers at the point of care [2-10]. These GL-based *Decision Support Systems* (DSSs) usually use a centralized infrastructure: the medical knowledge of the GLs is stored and retrieved from a central Knowledge Base (KB) library (examples of knowledge items stored in the KB are the medical definition of "High BP" or definition of a plan to take a specific medication). The patient data are stored and retrieved from an *Electronic Medical Record* (EMR). Finally, the centralized DSS engine applies the medical knowledge to the data to provide alerts and recommendations at the point of care to care providers (doctors or nurses), to the patient (for example through messages sent to his mobile), or even to Knowledge Engineers (KEs), to debug or simulate the DSS engine [11].

Traditionally, the GL recommendations generated by the centralized DSS engine are addressed to the care providers, such as the physicians at the point of care, rather to the patients at home. However, today the role of the patient in the process of care is becoming more common, and GL recommendations should also address patient behavior, especially in the case of chronic illnesses, where treatment must be continued outside the hospital and partly managed by the patients themselves or by his family members. Therefore, patients, and in particular, chronic patients, should be empowered with a *remote, local decision support assistance*, residing in the centralized DSS engine, to manage their own disease by extending both the GLs and the GL-based decision-support frameworks to provide guidance for patients outside the standard clinical settings. Thus, *a distributed decision support system* that includes patient guidance through the use of applications running on their mobile devices could provide the patients themselves with appropriate GL-based alerts and personalized recommendations, and could also monitor and react to changes in the patient's personal environment.

## 1.1 Motivation

### 1.1.1 The need for Distributed Decision Support Systems

The need for distributed DSS can be easily grasped through a brief look at the evolution of distributed computing architectures over the past three decades (Figure 1). It starts with centralized architectures based on *Mainframe computers* [12], and passes through distributed *Client-Server* architectures [13], which perform some of the computations on the client side, *Peer to Peer* architectures [14] in which two clients can directly interact, and *Mobile-Cloud* architectures, including the use of more robust architectures for Grid Computing [15] such as Hadoop [16], and a wide range of mobile-cloud applications [17].

Despite the relatively accelerated progress in the distributed computing area, the evolution of the area of distributed DSS in general, and specifically clinical distributed DSS, was rather slow, although it was increasingly discussed during the past decade: for example, in the report "*A Roadmap for National Action on Clinical Decision Support*" [18], prepared by Osheroff et al. for the 2006 AMIA conference, one of the strategic recommendations is *to collect, organize*, and *distribute* clinical knowledge with clinical DSSs. In a review of several types of intelligent DSSs from 2008, Zhou et al. defined the area of *distributed DSSs* as one of the new areas for research [19]. Also, the creation of an architecture for sharing executable DSS modules and services was mentioned by Sittig et al. in 2008 as one of the ten grand challenges for clinical DSSs [20]. This kind of architecture will support dissemination of the DSS knowledge to greatly speed the transition from research finding to widespread practice. However, when looking back on the past decade, only a few studies focused on the implementation of distributed DSSs, and even fewer of them focused on the healthcare domain.

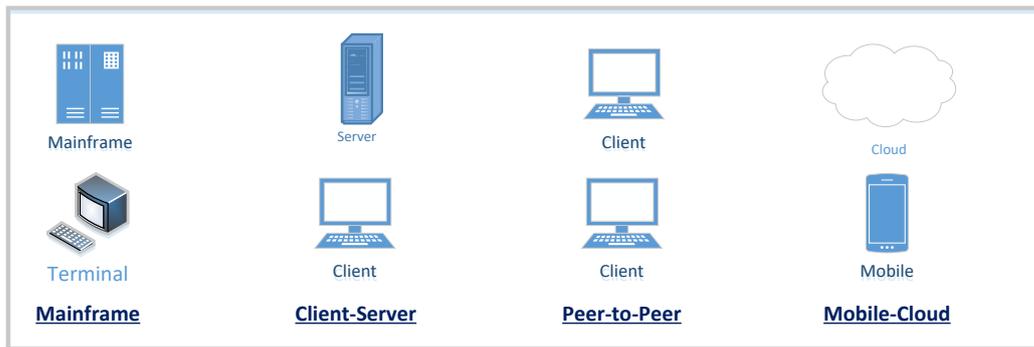

**Figure 1.** Evolution of distributed computing over the past three decades, starting with Mainframe computers, through Client-Server, Peer to Peer, and Mobile-Cloud architectures.

### 1.1.2 Dimensions of Distribution

Gachet et al. [21,22] define distributed DSSs as a "*collection of services that are organized in a dynamic, self-managed, and self-healing federation of hardware and software entities working cooperatively for supporting the solutions of semi-structured problems involving the contributions of several actors, for improved decision-making*". They also outline five main characteristics for distributed DSSs: 1) Computers are dispersed throughout a single organization or several organizations; 2) Computers are connected by a data communication system; 3) A common database is shared by all, but additional databases may exist; 4) All computers are centrally coordinated with an information resource management plan; and 5) Input and output operations are done within user departments.

However, these characteristics are defined in the context of *organizational DSSs* and focus on the flow of the data through the different services. Therefore, based on our experience from projects (such as the MobiGuide project, which we describe in detail in section 1.2), designing a *distributed* clinical DSS should also include additional distributed characteristics, such as:

- *Data distribution* – data might arrive from different sources (e.g., patient and local hospital)
- *Knowledge distribution* – the knowledge needed for the DSSs might can be saved in several knowledge bases (e.g., at patient mobile or at server)
- *DSS engine distribution* – the DSS engine(s) might be activated on different computational units (e.g., on the patient's mobile device or on a main server)
- *DSS recovery and robustness* – the distributed DSS framework should always preserve its integrity, maintain the continuity of the decision-support process, and handle failures at the different components (e.g., on the patient's mobile device, or on a main server).

Interestingly, very few studies appear in the literature regarding distributed DSSs in the healthcare domain. Most of the studies used distributed DSSs in different contexts, for example, a study on security protocol for clinical DSS to handle the ethical and privacy aspects of the data [23]. Other studies claim to have developed distributed DSSs; however, in the implementation of their architecture, the researchers often use a central server for both knowledge representation and for application of the DSS engine. An example is a system that provides decision support in the radiology domain for diagnostic purposes, using three main services: an information service for patient records, an analysis service that includes all the classification engines, and a radiologist interface service, all services connected through the Web [24]. Another example is a system supporting acute care within an ambulance by integrating several data sources, such as the electronic medical record (EMR) or the ECG, and processing them using a central decision-making server [25]. In another study, the authors hosted the knowledge base and the decision support engine in a community cloud, and allowed several hospitals to access the DSS centralized server, sending to the cloud DSS a small subset of their local EMR, to get reminders and alerts [26]. Thus, the *data* were distributed, but the *knowledge* and *processing* were centralized. Another study offers only a conceptual architecture for distributed DSSs, but again with separated central knowledge and DSS servers [27]. The only framework that handles all of the dimensions we mentioned, albeit in a simplified manner, is the TIDDM EU project [28-30] in the diabetes domain. In that study, the DSS was distributed between the patients at home using a Patient Unit (PU), and the physicians at the hospital using a Medical Unit (MU). The PU only performed simple calculations, such as computing the first level of simple decision tables and/or applying simple algorithms that suggest the next insulin dose depending on the current measurement, or a more complex level of computation, to detect in advance deviations from the expected metabolic control target. This computation is used to generate alarms and advice to the patients, and to trigger a connection with the MU. The programs of the PU are predetermined and cannot change. Although as we shall see, our proposed architecture is quite different, we consider the TIDDM architecture as a very early precursor of our highly generalized distributed architecture.

### 1.1.3 Our Specific Objectives and New Contributions

In this study, our overall objectives were to develop, and then to demonstrate in real life, the technical and clinical feasibility, of a new architecture for a distributed medical DSS, which provides patients, and in particular chronic patients, with continuous (i.e., 24/7) evidence-based, personalized guidance, offered through applications running on the patients' mobile devices, monitoring and reacting to changes in the patient's personal environment, and providing the patients with appropriate GL-based alerts and personalized recommendations; and increase the overall robustness of the distributed application of the GL.

As we shall see when we describe our methods, in the new architecture that we have designed, implemented, and evaluated in two different EU countries, whose objective is to personalize GL-based management of chronic patients over time, the algorithms themselves are dynamic, moving between the central server and the patient units. Thus, *the algorithmic contents of each patient's unit change over time*. In addition, as we shall see, the distribution of responsibilities between the central and local (patient) units follows certain well-defined guidelines. In the approach we describe in this paper, the focus is mainly on the *personalization of the computation* performed by the personal device of each patient, based on the personal data of that patient, and on the specific knowledge dynamically sent to that device from the central server (dependent on the patient's data), and *not* just on increasing the efficiency of the computation by using multiple processors, as is usually the case for most distributed-computation frameworks.

Thus, as we shall show in detail, the temporal data-based patterns that govern *when* each personal patient-associated mobile device calls the central computer for further assistance are customized to each guideline and patient-specific context; and so are the context-sensitive guideline segments that are sent, in response to the request, from the central server to each specific patient-associated mobile device. This conceptual framework is very different from splitting a large process into segments and running it on a multiple separate, but essentially identical, local clients.

We have developed and implemented our new architecture for distributed GL-based DSS within the *MobiGuide* EU project. We assessed the new GL specification process, by specifying two complex guidelines: for Gestational Diabetes Mellitus, and for Atrial Fibrillation. We have then evaluated the new computational architecture by applying the two GLs to two different groups of gestational-diabetes and atrial fibrillation patients, in Spain and Italy, respectively.

Thus, our work demonstrates the immense importance of effective procedural medical knowledge engineering, geared ahead of time to a distributed, large-scale application mode, and the advantages of *a practical application of distributed artificial intelligence techniques in the real clinical world*.

### 1.1.4 The role of mobile devices in distributed DSSs

These days, a distributed DSSs for patients is most likely to involve a mobile device, a trend recognized long ago [31, 32]. A systematic review of 111 studies comparing telemedicine with usual care for adults with diabetes found that telemedicine can be efficient for monitoring and controlling the HbA1C level; however, there was not enough evidence that patient outcome also improves due to this technology [33]. Several research and review studies found gaps between GL-based recommendations and the functionality offered by current mobile applications [34, 35]. Connecting the mobile device to the EMR seems crucial, as concluded by a systematic review of 192 mobile applications using a clinical DSS, to reduce the time the user requires to interact with the system [36], and to exploit the full impact of a mobile-based system [37]. An integration with the EMR is also important to enhance personalization. In our view, recommendations should be personalized not just clinically but also in the sense of considering the patient's personal schedule, important external events, and personal preferences corresponding to changing contexts, including non-clinical contexts that were not accounted for in the original GL, such as the patient living alone, or the battery status of the mobile device.

To the best of our knowledge, *there is currently no general designed, implemented, and evaluated framework for a personalized distributed medical DSS.*

Having described in our objectives the desiderata for what we consider a truly distributed DSS, and the gaps in the current architectures, we now describe the architecture we have designed, implemented, and evaluated, for real-time distributed decision support in the context of patient guidance systems, the European Union's (EU) MobiGuide project [38-40], and the several innovations that it encompasses.

Understanding the MobiGuide project's aims and methods will facilitate the presentation of our new distributed-application methodology, the Projection/Call-Back methodology, and the details of its rigorous multi-national evaluation. Thus, we start by briefly reviewing the MobiGuide project, before moving on to our proposed methodology and to its implementation and evaluation.

## 1.2 The MobiGuide Project

The main goal of the MobiGuide project is to develop a distributed patient guidance system that integrates historical hospital records and current monitoring data into a unified PHR accessible by patients and physicians anytime, anywhere, and that provides personalized, secure, clinical-guideline-based guidance both inside and outside standard clinical environments [38].

Figure 2 provides an overview of the MobiGuide ubiquitous system: Patients wear a *Body Area Network* (BAN) that includes biosensors, such as an ECG belt or a blood-glucose monitor [41], and which sends the signals produced by the sensors to the patient's smartphone. Clinical practice guidelines are stored in a Knowledge Base. The knowledge base includes GLs that contain detailed clinical management instructions (therapy plans). It also includes the definitions of several potential sub-contexts within the GL (e.g., the context of an irregular meal schedule) that might affect the medical plans that are recommended by the GL. One innovation in the MobiGuide context is that they can be *personalized*; for example, the predetermined context of an irregular meal schedule might be *induced*, for a particular patient, by the event "vacation", which she can add to her personal context list, and which creates in her record the predetermined context [38, 39].

The main novelty in MobiGuide compared to traditional GL-based decision-support systems is that the *decision-support system is distributed into a centralized decision-support system at the back-end, and a set of multiple local mobile-based DSSs*. The management plans applied by each of the local DSSs can change dynamically, on a patient-by-patient basis, as we shall see, coordinated by local events and by the central DSS.

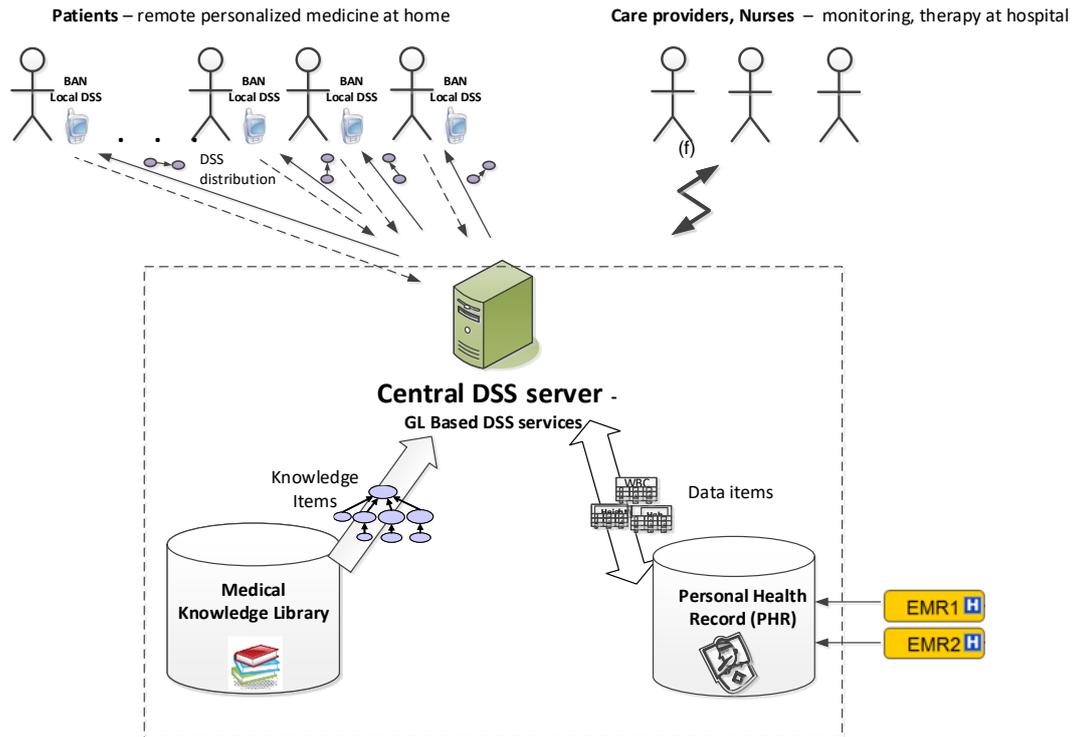

**Figure 2.** A high level overview of the MobiGuide Ubiquitous Guidance System. The decision-support system (DSS) is at the center; data storage is performed by the Personal Health Record (PHR), which acquires data from the hospitals' electronic medical records (EMRs), the patient wearable Body Area Network (BAN) sensors, the patient's smartphone, and the DSS.

The backend, *central* decision-support system, provides full decision support based on all patient data available at the personal health record and on the full GLs representation, available in the knowledge base. Care providers can access the *central* DSS using their work computers.

Patients can access the *local* decision-support system on their mobile device through their smartphone interface. The *local* DSS includes a patient's user interface as well as signal-analysis and decision-support algorithms that run on the processor of the smartphone, and therefore has access, with respect to knowledge, only to a limited set of relevant GLs (which, as we shall see, are in fact a distributed portion of the main GL, assigned for application to the mobile device), and to a limited set of data (in particular, not the historical EMR). The PHR of the MobiGuide system integrates different data sources, such as data from EMRs of several hospitals (e.g., hemoglobin A1C values, current medication prescriptions), sensor data collected via the BAN, patient input collected via the smartphone, and decision-support recommendations provided by the DSS (e.g., a recommendation to decrease carbohydrate intake). At the heart of the central DSS sits the PICARD DSS engine [11, 42] as the backbone architecture, which runs the main GL, sending the patients personalized alerts as necessary, and providing context-sensitive guideline-based recommendations to their care providers.

The MobiGuide architecture was tested rigorously through clinical pilots performed in two different domains: 1) *Gestational Diabetes Mellitus* (GDM), with or without hypertension, and 2) *Atrial Fibrillation* (AF), demonstrating its context-sensitive, personalized nature [39] and its potential clinical value and enhancement of patient satisfaction [40].

### 1.2.1  A Brief Preview: The Projection/Call-Back Methodology

The distributed model of the MobiGuide framework was not implemented as a regular service oriented architecture [43], which might be more suitable for distributing a process inside a hospital. Instead, we have chosen to split the architecture into two main components: a *back-end Decision Support System* (*BE-DSS*) residing on a server system (this could be a cloud server, or, as in our case, on premise servers in hospitals), and a *mobile DSS* (*mDSS*) residing on the patient's mobile device. The local mDSS is necessary, as will be explained in the next section, to distribute computationally intensive monitoring and decision-making processes, with respect to data and knowledge requirements, to the local device level.

Thus, to accommodate the need for both central and local decision making in the MobiGuide project, we decided to design, implement, and assess a new architecture for dynamically distributing evidence-based decision support, which answers the core need for patient-centered, guideline-based management of chronic patients.

Our solution involves dynamically *projecting* computational tasks from the central server to the local (mobile) computational modules, and *calling back* the central server from a local (mobile) device when a centrally available resource (e.g., knowledge) is needed. Thus, we refer to the methodology underlying our dynamic distributed decision-support architecture as the ***Projection/Call-Back*** (***PCB***) distributed-application methodology.

In Section 2, we describe our PCB methodology in detail, and, for each aspect of it, describe its implementation.

In Section 3, we describe our evaluation methodology, which was designed to answer the following two main research questions:

1) Is the specification of the knowledge embodied in the distributed guidelines, using the new PCB methodology, functionally feasible?
   To assess the answer to this question, we have specified, using the new methodology, two very different guidelines, with respect to their character and frequency of monitoring and application actions (management of *Gestational Diabetes Mellitus* (GDM) in pregnant women, and management of *atrial fibrillation* (AF), a form of cardiac arrhythmia). We also measured the number of each type of resultant knowledge roles in the knowledge base representing the specified guidelines.
2) Is the implementation of the PCB methodology functionally feasible (i.e., for *real* guidelines applied to *real* patients in *real* time)?

   To assess the answer to this question, we applied the two guidelines within the full-fledged MobiGuide architecture, managing gestational diabetes patients in Barcelona, Spain, and Atrial Fibrillation patients in Pavia, Italy. We also documented the full interaction between the BE-DSS and the mDSS, and measured how many projections and callbacks were made, and when.

In Section 4, we describe our evaluation's results. We summarize and discuss our methodology and its results, and the conclusions we can draw from them, in Section 5.

## 2 The Dynamic Distributed Decision-Support Methodology

When implementing distributed decision support, multiple considerations characterize the decision processes that are best performed at the central, BE-DSS level, versus those that are best performed locally, at the mDSS level. To better understand our methodology, we start by examining more closely, exactly which decisions need to be made centrally, and which are better left to a local computational module.

### 2.1 The characteristics that affect the decision regarding central versus local computation

Consider the text in Figure 3, which describes the *ketonuria management plan*, taken from the GDM guideline.

> *Measure ketonuria level daily. If no positive ketonuria was detected for two weeks, measurement frequency can be reduced to twice a week. If there are two positive ketonuria values within one week, ask the patient whether she ate enough carbohydrates. If this is the first time that the patient had two positive ketonuria values since the last visit to the GDM clinic, suggest to the patient to increase the amount of carbohydrates during dinner.*

**Figure 3**. The text of the ketonuria management plan, taken from the *Gestational Diabetes Mellitus* (*GDM*) guideline.

In terms of DSS distribution, it seems that most of the computation effort can be done at the local level by the mDSS: ketonuria monitoring (both daily and twice a week schedules), detecting positive/negative values within a week or two weeks, and asking the patient about his diet can be handled by the mDSS even in an offline mode, rather than by the BE-DSS, which does not need to perform continuous monitoring of the schedule of all of the patients. This can save a large amount of back-end computation. On the other hand, switching between the two schedules, and evaluating the query of whether the unbalanced occurrence was the first time for the patient, is better handled by the server as the BE-DSS because it has access to the full GL and the PHR, which is essential for making the decision.

Another example for distribution choice is using the mDSS as resource for intensive computation of short-term patterns such as "AF episode" that require only local data, and therefore should be distributed. However, it is limited in storage space and memory, and cannot access the PHR to retrieve data. Instead, it can make analysis only on data of a particular patient, which comes from the mobile sensors. In this way, the mDSS can provide personalized DSS (e.g., delivering a reminder 30 minutes before the breakfast time declared for a particular patient) to the patient. On the other hand, at the central level, the BE-DSS is responsible to make recommendations personalized by accessing the PHR and elaborating each recommendation with user preferences. Thus, the BE-DSS is responsible to create and deliver the personalized recommendations, but invoking them and managing them at the mobile is the sole responsibility of the mDSS. Another responsibility of the central level back-end might be to calculate complex temporal patterns that require a longitudinal record and possibly additional sources of knowledge, or of historical data from PHR, such as a pattern of "repeating episodes of AF" with certain temporal and statistical characteristics.

Table 1 summarizes the characteristics we found based on our experience in the project of the local (mDSS) versus the central (BE-DSS) computational decision processes. Also in the table are several examples of the distribution of decision making between the BE-DSS and the mDSS for the GDM and AF GLs. These characteristics can be viewed as a meta-ontology of decision processes for the knowledge engineer, suggesting, typically during knowledge specification time, when a certain decision task should be delegated to the local mDSS, and when it should best be left to the central BE-DSS. For example, in the GL elicitation process, these characteristics might help to decide whether to implement a decision process locally, i.e., to project to the mobile device, from the central BE-DSS, the necessary procedural and declarative knowledge. Then, they might enable the evaluation of the decision together with the local data available, and to compute the result of the decision process. In the next section, we explain in more detail what this projection process entails.

**Table 1.** Characteristics of the local *mobile decision-support system* (mDSS) versus the central *back-end decision-support system* (BE-DSS) computational decision processes.

| Characteristic | mDSS level justification | | BE-DSS level justification | |
|---|---|---|---|---|
| | **Description** | **Examples** | **Description** | **Examples** |
| Computation | Distributed application of personal, intensive computation of short-term patterns that require only local data | Calculation of "Atrial Fibrillation (AF) episode" or "pattern for monitoring two weeks of normal blood-glucose" | Central application of complex temporal patterns that require a longitudinal record and possibly additional sources of knowledge, or of historical data | Calculation of the pattern "repeating episodes of AF" with certain temporal and statistical characteristics |
| Knowledge | Partial knowledge; typically, the current guideline's declarative knowledge, and knowledge of several context-sensitive procedures that are relevant to the current GL application session | "Monitoring BG" procedural plan, or "Abnormal BG" declarative definition | Full access to all knowledge during GL application session, including knowledge that requires switching to another sub-plan of the GL, or even to another GL | The complete GDM or AF GL |
| Data required or available | Decisions that are based on recent, short-term data | "Monitor last week of heart rate", or "Monitor today's BG measurements" | Decisions that are based on historical, long-term data of the full historical patient record | "Monitor AF episodes of the previous several hours or days or weeks", or "GDM during previous pregnancies", or "Count last visits of patient" |
| Horizon of future Recommendations | Short-term horizon | "check blood glucose now" or "measure urine for ketonuria today" or "visit your dietician" | Long-term future horizon | "Visit a dietician every month for the next four months" |
| Subject of Data | Data of patient only | Specific values of BG, BP of patient | Population data of all patients is needed for decision | The average BG counts at a clinic |
| Data sources | Decision relies only on data entered from local sensors or entered manually by patients | Diet non-compliance times manually entered entry, or BG entered automatically coming from sensors | Decision relies on data from different sources such as EMR data containing laboratory tests, medication prescriptions, and diagnoses | Does the Ketonuria abstraction have a negative value over the past week AND has the Diet not been changed since the last visit? |
| PHR access | Cannot access the PHR to retrieve additional data, and cannot retrieve special types of data such as previous abstractions and recommendations | | Can access the PHR to query for any type of EMR data, monitored data, previous abstractions of the data, and previous recommendations | Is the patient eligible for the "Pill-in–the-pocket" emergency plan? |
| Personalization | Invoke the personalized recommendations that are sent to the patient | Setting personal times for reminders of BG | Initially generates recommendation from the GL, and then personalizes them by elaborating them with user preferences from the PHR, if there are any. | Setting default times for recommendations |
| Quality of data | Quality of acquired data can be assessed locally using only mobile device resources | Checking a locally measured blood pressure value | Assessment of quality of data requires referring to data from the PHR, or requires knowledge, computation, or other resources that are available only to the BE-DSS | Assessing the likelihood of a temporal pattern of weight measurements that requires context-sensitive temporal-abstraction knowledge |

## 2.2 The Projection/Call-Back Distributed-Application Methodology

Having better understood the considerations that led us to outsourcing a computation to a local device, versus performing it centrally, we now focus on our specific methodology for solving the problem of dynamically distributing the computational tasks involved in evidence-based decision support to a large number of patients: the *projection/call-back* (*PCB*) methodology.

In the PCB methodology, we *project* context-sensitive knowledge from the central server to a local device, such as the patient's smart phone, and *call back* the central server from the local device when the local device requires additional, centrally accessible resources to make a decision. As we shall see, in response to such a call-back request, the central, backend server might well send the local device a different projection.

### 2.2.1 An overview of the knowledge projection methodology

As explained in section 2.1, deciding where decisions and actions should be applied (at the BE-DSS level or at the mDSS level), has several characteristics and considerations that must be addressed by the knowledge engineer and expert physicians during the knowledge specification phase. However, once it is decided that the decision can be applied by the mDSS, i.e., can be answered by the patient when he/she is not with the care provider, the mDSS needs some knowledge about the GL, i.e., there is a need to project ("download") small portions of the GL, referred to as *projections,* from the BE-DSS to the mDSS in the mobile device, which applies it using the mobile local resources. As this projection relates to knowledge and not to data, it is called *knowledge projection,* or "projection" in short. Each portion in the GL, tagged to be projected is called a "projection point". Only parts of the GL that are applicable to the current state of the patient are projected. Hence, the projected knowledge also includes knowledge about the current state of the patient.

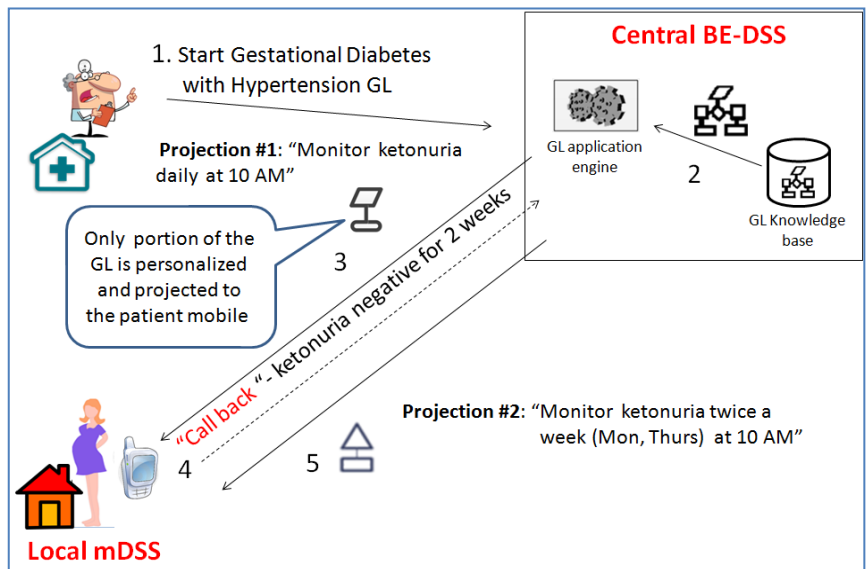

**Figure 4.** An example of a projection/call-back model workflow.

Figure 4 demonstrates the MobiGuide PCB model in the context of the text described in Figure 3: After the physician initiates application of the Gestational Diabetes Mellitus (GDM) guideline (number 1), the BE-DSS retrieves the full GDM GL from the GL knowledge-base (KB) server (number 2), but sends only the specific sub-plan "monitor ketonuria daily", which was previously tagged as a projected plan, to the mDSS; this sub-plan (representing the current treatment plan for that individual, and personalized with patient preferences and current context) is then applied by the mobile device (number 3).

At any time, a certain predefined *breakout* temporal pattern might be detected by the mDSS (as part of the projected sub-plan) or by the BE-DSS. When a breakout pattern is detected by the mDSS, the mobile device sends a message to the BE-DSS to take control of the current GL application thread, indicating which breakout pattern was triggered. This message to the BE-DSS from the mDSS is called a *callback*, and is also predefined within the projection (number 4). A callback, or a detection of such a pattern by the BE-DSS, will in some circumstances lead to the sending of a new projection file by the BE-DSS to the mDSS (number 5) (e.g., effecting a change in the treatment plan). For example, the breakout pattern "two weeks of negative ketonuria" detected by the mDSS (or by the BE-DSS, depending on how the pattern was defined: as a breakout pattern, for the mDSS, or as a monitored pattern, for the BE-DSS) might lead to stopping the "twice a week ketonuria measurement" sub-plan, and starting a new "ketonuria measurement" sub-plan. In both cases, the switch between the current sub-plan and the new sub-plan occurs at the BE-DSS level, which sends a new projection file to the mDSS. A projection file includes one or more projected plans, which will be activated immediately and applied by the mobile device.

## 2.2.2 The Distribution policy

In Section 2.1, we described several characteristics of the local (mDSS) versus the central (BE-DSS) computational decision processes. These characteristics help us in deciding when a certain decision task should be delegated to the local mDSS, and when it should best be left to the central BE-DSS. Applying a more general perspective, we noted that there is a need to define explicitly the "Distribution policy" that determines the overall policy for splitting the computational burden between the mDSS and the BE-DSS. For example, in one extreme case, when no mDSS is available on the mobile device, the GL might be run exclusively by the BE-DSS; the BE-DSS sends the mobile device's Application Interface (API) a direct message containing the recommendation's details, including a potential interaction with the patient. This is bypassing the projection and callback mechanisms. At the other extreme, the GL might be fully projected to the mDSS, which runs it completely locally, without connecting to the BE-DSS. In between, there are several distributed decision-support options in which both the mDSS and the BE-DSS are used. For example, the BE-DSS might be running the GL in a "*full-shadowing*" mode, in which it runs concurrently with the mDSS, silently following every action performed by the mDSS, in case the mDSS encounters a technical problem (this mode can be viewed as a "hot backup" option). Alternatively, the BE-DSS might run the GL in a "semi-shadowing" mode, in which it only monitors the mDSS results (e.g., making sure that recommendations are sent to the patient or to the physicians). In both of the last cases there is some overhead, as the BE-DSS performs computational efforts that can be delegated to the mDSS.

Thus, the ideal option might be to switch control between the mDSS and the BE-DSS: When some of the GL needs to be projected, the BE-DSS suspends the GL application session, and "passes control" to the mDSS by sending it a new projection file, and only listens to a message from the mDSS to "take control" back to resume the application of the GL (see next section for more details). Table 2 summarizes the five types of "distribution policy" we identified, and suggests situations in which each of them might be used.

In the current study and during the implementation and evaluation of our dynamic distributed decision-support framework within the MobiGuide project, we implemented the most general policy, i.e., the "passing of control" policy, as the default behavior, since we wanted to better understand the implications of implementing and applying in real life several GLs through a fully distributed decision-support process.

**Table 2.** Distribution policy between the mobile decision-support system (mDSS) and the Back-End decision-support system (BE-DSS).

| Distribution policy | Description | Remarks |
|---|---|---|
| **Full mDSS** | GL is projected completely to the mDSS and run exclusively by the mDSS | • GL is very simple, and contains only one main path of treatment, with no need to project new GL portion from server<br>• The mDSS is available at extreme high service level and has high computation capacity<br>• No need to make decisions about multiple patients |
| "Full shadowing" | BE-DSS completely monitors mDSS actions | • When mDSS might be not available or cannot be relayed to have high service level<br>• The BE-DSS is running in parallel to the mDSS to make sure alerts are sent to the mobile |
| "Semi-shadowing" | BE-DSS monitors mDSS results | • mDSS cannot make complex calculations (e.g., calculate AF episodes) |
| "Passing of control" | mDSS/BE-DSS alternately switch control | • True distribution – no redundancy in computation; in each time only one component manages the therapy |
| **Full BE-DSS** | run exclusively by the BE-DSS | • No mobile at all – all the GL is running and monitored by the back-end DSS |

### 2.2.3 Specifying the Guideline in the Terms of a Distributed DSS

Specifying a GL for application by a fully distributed DSS requires a different strategy from traditional GL knowledge engineering. As explained above, it involves multiple new challenges, such as: deciding at which level (mDSS or BE-DSS) each plan or decision should be placed, deciding which action or plan needs to be performed, and deciding which breakout patterns should trigger callbacks to the BE-DSS. This process is performed during the GL specification phase by a knowledge engineer in collaboration with expert physicians, as part of the process of creating a consensus regarding the GL [44].

The main challenge in the projection process is to choose at which level (mDSS or BE-DSS) the decisions should be placed. When choosing these projection-points in the GL, consideration should rely not only on technical analysis methodologies, but also consider clinical properties, such as whether the decision could be taken by the patient alone (in this case it could be delegated to the mDSS) or whether it should be done by the care provider (hence should be done on the BE-DSS). Therefore, to start, to identify and tag the projection points, we used the distribution decision support, described in Figure 5: The first phase in this methodology is to make local consensus of the GL (Rectangle 1). The local consensus is a structured document that describes schematically the interpretation of the GL agreed upon by both the expert physicians and the knowledge engineers, and includes the clinical directives of the GL and the semantic logic of the specification language [44]. After creating the local consensus, the elicitation phase splits into two parallel branches. The first is the more "traditional" workflow, and is intended for the health-care professional (Rectangle 2). The second (Rectangle 3) includes the parallel part of the process, which focuses on the *patient's* behavior, and which is added to the GL during the GL elicitation and representation phase. We call this model the "Parallel Workflow" model [45]. Each parallel workflow, i.e., a part of the GL that is intended for the patient, is a ***potential candidate for projection***.

During the "parallel workflow" process of determining whether tasks should be delegated to the local device, the knowledge engineer or the expert physician might use the characteristics listed in Table 1. Finally, for each decision that can be taken by the patient, a projection model is built by the knowledge engineer using the knowledge acquisition tool in the terms of projecting it to the mobile (Rectangle 4). Figure 6 shows an example of "parallel workflow" for "monitoring the blood glucose" plan for the GDM GL. This process was tagged by the expert physician as "can be handled by the patient", and therefore was also tagged as projection point that can be downloaded to the mDSS and applied at the mobile.

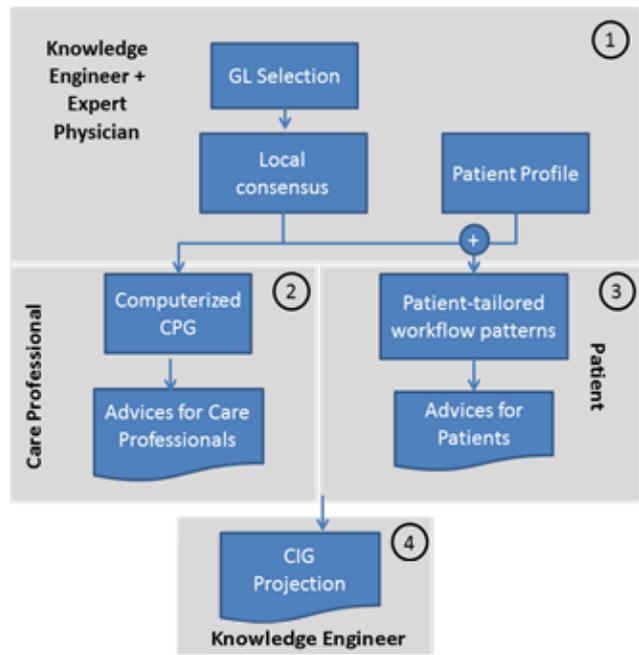

**Figure 5.** The methodology for guideline formalization and patient-tailored workflow patterns identification that produces a customized guideline.

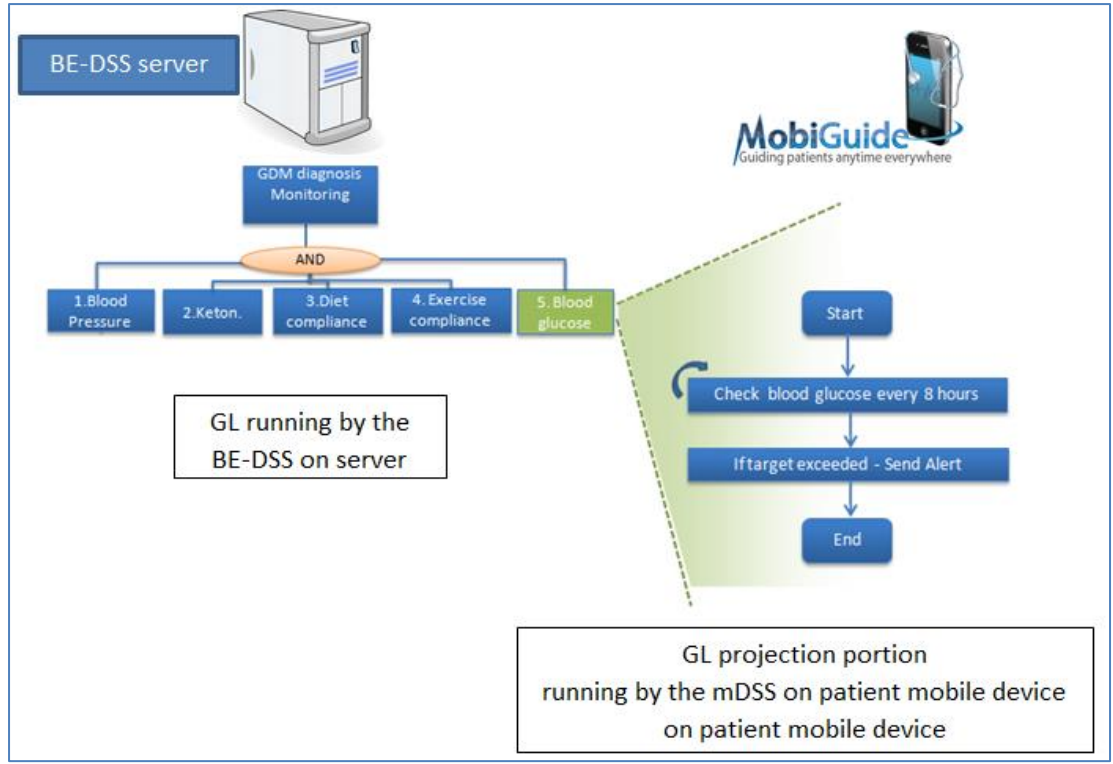

**Figure 6**. An example of "parallel workflow" for the "monitoring blood glucose" sub-plan, and its tagging as projection point.

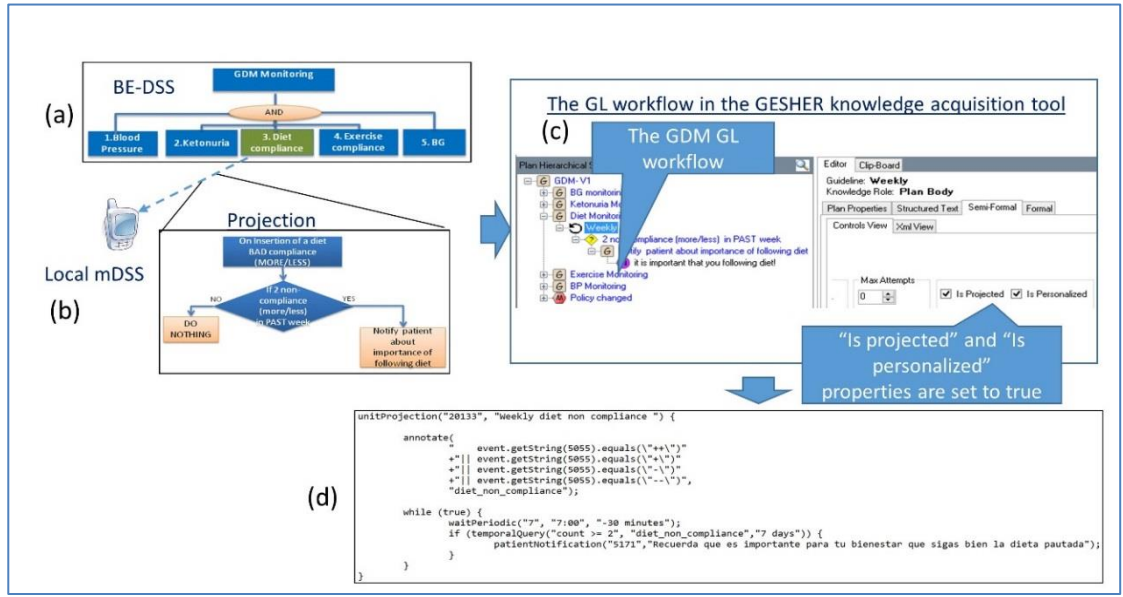

**Figure 7.** An example of a parallel workflow and projection points in the *gestational diabetes* (*GDM*) guideline.

Figure 7 shows an example of a parallel workflow and projection points in the GDM guideline: a) At the BE-DSS, several plans are indicated in parallel, as specified in the "traditional" guideline. b) The "diet compliance" plan is specified as a parallel workflow for the patients. c) The knowledge engineer uses the GESHER knowledge acquisition tool [46] to explicitly specify projection points, by setting the "is projected" property of the (sub)plan to "true". Note that she also indicated that this plan needs to be personalized by setting the "is personalized" property to "true". d) The projection file, customized to the patient, is ready to be applied by the mDSS.

Figure 8 shows how we implemented the tagging of plans as plans to be projected as part of the GL, using the GESHER knowledge acquisition tool [46], as part of the *Digital electronic Guideline Library* (DeGeL) [47], in the case of the GDM GL: At specification time, the knowledge engineer checked the "*is-projected*" property of the sub-plans that were determined as sub-plans that need to run at the mDSS level, in this case the "monitor ketonuria daily" sub-plan (number 1) and the "monitor [for ketonuria] twice a week" sub-plan (number 2). Note that in both cases, two sub-plans are tagged as *projected*: the first is a periodic sub-plan for measuring the ketonuria each day (the circular arrow shape); the second is a monitoring sub-plan (the hexagonal shape), which in fact monitors for a breakout pattern (in this case, the pattern "two positive values of ketonuria in a week"), and which, if detected, asks the patient a question regarding their diet and triggers a callback to the BE-DSS to determine how to proceed. Note that several projected plans (depending on their internal eligibility criteria and the GL's overall workflow) might be sent in the same projection file to the mobile device. The basic language used for representing the GL, underlying the GESHER GL-specification tool, is Asbru [48], and its hybrid-Asbru extensions [49]; we have augmented them using the projection and callback tags.

Regarding other GL application frameworks, to the best of our knowledge, only the GLARE GL application framework [50] explicitly introduced the concept of distributed GL-based decision support, implemented by managing several agents that interact in different clinical settings, called "contexts". However, that extension was intended to deal mostly with human interaction and communication, and with human resources management; the agents were human; and none of the agents mentioned was the patient.

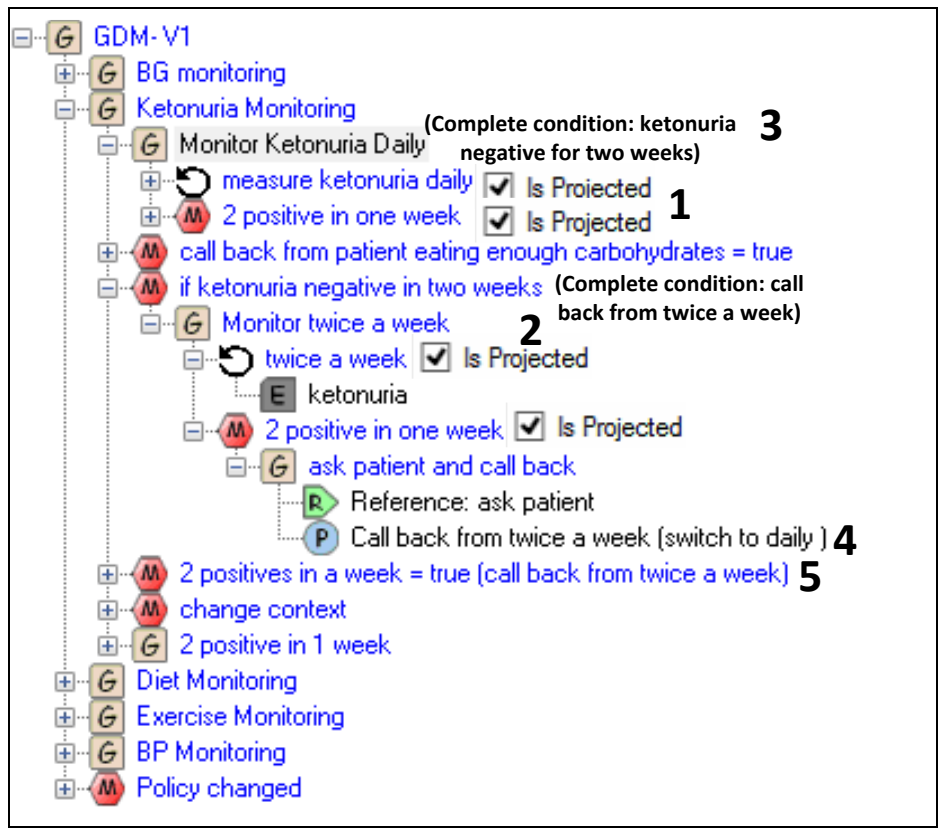

**Figure 8**. The GESHER interface and the tagging of projected plans in the guideline, in the case of the *Gestational Diabetes* (*GDM*) guideline.

### 2.2.4 Definition of appropriate callbacks

As explained above, projections and callbacks support a continuous dialog between the BE-DSS and the mDSS. Thus, the GL created to support a distributed DSS is also specified in terms of messages between the mDSS and the BE-DSS. This includes specifying the projections to send to the mobile device, the breakout patterns to be detected by the mDSS, and the associated callbacks from the mDSS to the BE-DSS. On the other hand, at the BE-DSS level, monitoring plans are "listening" to all of the relevant breakout patterns, and to callback messages coming from the mDSS, which might cause the BE-DSS to send the mDSS a message to stop an existing projected plan, or to send a new projected plan to be activated.

The implementation of this unique dialog in terms of GL specification is shown also in Figure 8. First, "monitor ketonuria daily" is projected to the mobile (number 1), and accordingly a monitoring sub-plan to detect the temporal pattern "ketonuria has been negative for two weeks" is activated, in this case, at the BE-DSS level, as determined by the knowledge engineer who built the parallel workflow for this section of the GL (note that in this particular case, it could also be applied by the mDSS by projecting it to the mobile device). When that sub-plan is triggered (by another part of the BE-DSS, an *intelligent monitoring* module that monitors the data for knowledge-based temporal patterns and that is subscribed to the urine measurements reaching the PHR server; see section 3, number 2), two events occur: 1) The *complete condition* (in the terms of the Asbru language [48], in which MobiGuide GLs are specified) for the daily monitoring sub-plan is triggered, thus causing the BE-DSS to send a projection to the mDSS to stop it (number 3), and 2) a new sub-plan, to reduce the frequency of monitoring to twice a week is started and is projected by the BE-DSS to the mDSS.

Note that the new projected "monitor ketonuria twice a week" sub-plan, which replaces the originally projected "measure ketonuria daily" sub-plan, includes a call-back instruction (number 4) to the BE-DSS in case the mDSS detects the breakout pattern of two positive values of ketonuria in a week. When the mDSS detects this breakout pattern, a callback is sent to the BE-DSS. This call-back is constantly monitored (through the intelligent monitoring module) by a specific monitoring sub-plan (number 5); thus, when the callback arrives at the BE-DSS, it causes the BE-DSS to send a stop message regarding the sub-plan for twice weekly monitoring (number 2), and to project to the mDSS a daily monitoring plan.

## 2.3 The projection engine

Recall that at the heart of the MobiGuide and other similar evidence-based management projects sits the PICARD DSS GL-application engine [11, 42]. To support the distributed DSS projection model, we developed a new component within PICARD DSS, the *projection engine*, which extends the functionality of our existing GL application engine. The extended BE-DSS architecture is shown in Figure 9: the GL application engine gets the GL knowledge from the GL KB, and applies it. When the GL application engine finishes, the projection engine examines which parts of the existing activated sub-plans need to be projected. The projection engine then retrieves the preferences and personalized contexts [39] of the patients from the data integrator through the data and knowledge services layer, and may also perform queries via the intelligent monitoring module (e.g., to get the current context of the patient). Then, the projection engine generates the projection file, which is sent by the GL application engine to the mDSS through the *Body Area Network* (BAN) back-end server [41], which mediates between the BE-DSS and the mDSS.

The projection engine produces two types of projections: (1) *Context-inducing projections* listing personal (patient-specific) events that induce predefined customized contexts appearing in the GL, within which the guideline's actions might be modified; and (2) *Procedural projections* – including sub-plans for general treatments or for specific treatments relating to personalized-contexts.

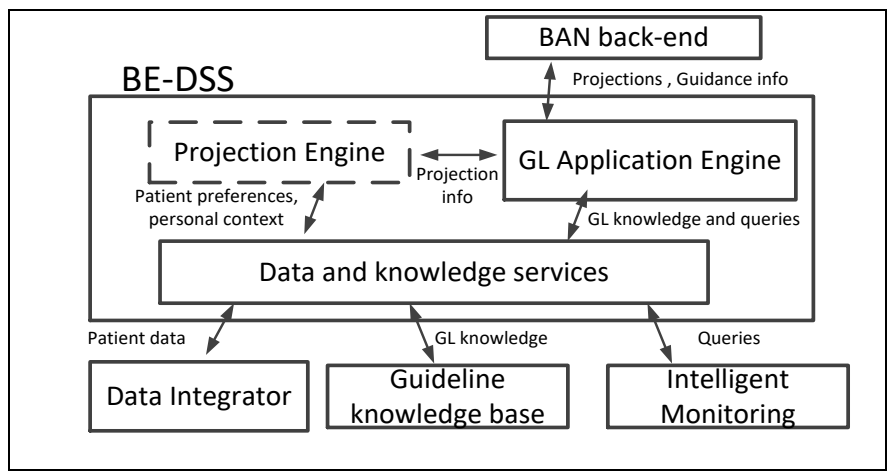

**Figure 9**: A high level architecture of the *Back-end Decision Support System* (*BE-DSS*) and of the projection engine.

### 2.3.1 Context-inducing-knowledge projections

One of the innovative features of the MobiGuide system is its support for personalization of evidence-based care. Personalization of the GL occurs when the patient is enrolled as a user of the MobiGuide system. In this step, the set of predefined clinical contexts are retrieved from the knowledge base and shown to the patient. These contexts are already part of the customized GL, i.e., a GL to which specific contexts were added to customize it to application through a mobile device, for management of ambulatory patients (e.g., a High Carbohydrate Meal context, or an Irregular Schedule context) [51]. The patient can then choose the corresponding personal event that induces each of these predefined contexts (e.g., a "Wedding" or a "Vacation" event, respectively). The patient's personal events induce the predefined customized contexts, in the same way that all other contexts are induced in the patient's data: In a dynamic fashion, as patient data arrive to the mobile device or to the PHR; the induced context starts and ends within a particular time span relative to the start and/or end of the inducing event, which is defined as part of the induction knowledge. Thus, we refer to the knowledge involved as *dynamic induction relations of contexts* (*DIRCs*) [51], a concept originally appearing in the dynamic temporal-abstraction contexts theory [52]. The mapping between the personal events and the contexts that they *induce*, i.e., the personal DIRCs, is stored in the PHR and is sent to the mDSS before starting the GL application session.

The mDSS uses a special context-inducing projection to send to the patient's interface the list of personal DIRCs, i.e., the personal events that were selected during the initial enrollment session, which predefined customized-GL contexts induced by each personal event, and when. For example, for a given patient, the "Vacation" or "Holiday" events might induce (immediately) the predefined context "irregular schedule", and a wedding event might induce the predefined context "High Carbohydrate Meal".

For example, a patient might register on her smartphone the personal event "at work"; the mDSS reports this event to the PHR. The event, only for her, induces (at the level of the BE-DSS) the "Regular-Schedule" context. If this patient subsequently reports the personal event "holiday", this will then be reported by the mDSS to the PHR, inducing, for her (at the level of the BE-DSS), the context "Irregular Schedule", thereby leading the BE-DSS to send the mobile device a second projection to be applied by the mDSS, with a different periodic monitoring rate. The BE-DSS always generates the projections to be sent to the mobile device according to the current (possibly induced by a personal event) context of the patient.

Figure 10 shows an example of the declarative knowledge. The declarative projection also includes the *Quality of Data* (QoD) [53] information used by the *Quality of Data Broker* located at the mobile, in order, for example, to "ignore" invalid input data (e.g., an out of range blood pressure value).

```
<Projection GLID="19857" GLName="GDM" CurrentContext="0">
    <QualityOfData>
        <QualityOfDataItem id="5560" description="Low" relateTo="4985,4986,4987,4988"/>
        <QualityOfDataItem id="5560" description="Very Low" relateTo="4985,4986,4987,4988"/>
        <QualityOfDataItem id="5559" description="Low" relateTo="5177,5178"/>
        <QualityOfDataItem id="5559" description="Very Low" relateTo="5177,5178"/>
    </QualityOfData>
    <personalEvents>
        <PersonalEvent ConceptID="5128" personalEvent="Diario">
            <Reminders>
                <Reminder Value="09:00" RemindTime="-5" GesherId="4985" Unit="minutes"/>
                <Reminder Value="10:00" RemindTime="-5" GesherId="4986" Unit="minutes"/>
                <Reminder Value="15:00" RemindTime="-5" GesherId="4987" Unit="minutes"/>
                <Reminder Value="22:00" RemindTime="-5" GesherId="4988" Unit="minutes"/>
            </Reminders>
        </PersonalEvent>
        <PersonalEvent ConceptID="5138" personalEvent="Festivo">
            <Reminders>
                <Reminder Value="10:00" RemindTime="-5" GesherId="4985" Unit="minutes"/>
                <Reminder Value="11:00" RemindTime="-5" GesherId="4986" Unit="minutes"/>
                <Reminder Value="15:00" RemindTime="-5" GesherId="4987" Unit="minutes"/>
                <Reminder Value="22:00" RemindTime="-5" GesherId="4988" Unit="minutes"/>
            </Reminders>
        </PersonalEvent>
    </personalEvents>
</Projection>
```

**Figure 10**. A context-inducing projection.

### 2.3.2 Procedural projections

Procedural projections are generated at run-time by the projection engine. The engine checks the "*is projected*" property for each sub-plan (see Figure 8), and if it is set to "true", the sub-plan's corresponding projection file is generated and added to the projection collection (see below for more details about projection collection). Otherwise, the BE-DSS engine continues to apply the sub-plan. Also, as part of the projection process, the specific thresholds of the patients (e.g., personal target exercise levels), and the preferences relating to the personal contexts of the patient are retrieved from the PHR and are set in the projections.

Each projection file is decomposed into several "unit-projections". Each unit-projection is a single sub-plan (see Figure 11). For example, the sub-plan "monitoring blood glucose once a week" is decomposed into two unit-projections: 1) the sub-plan for blood glucose measurement schedule and 2) the sub-plan to monitor several days of high fasting blood glucose levels, signifying that the patient is not well-controlled. Each projection and each unit within a projection starts independently, as soon as it arrives at the mobile device. Each unit has its own set of internal temporal constraints, including temporal relations among different actions. Thus, all important temporal-constraint knowledge resides (is encapsulated) within the projected units and not among them, hence concurrency problems are not an issue.

```
projection("19857", id="184");
stop("20091,20092");
start("20102,20130");

unitProjection("20102","Semi-Routine Daily BG Fasting measurement") {
        while (true) {
                waitPeriodic("1,2,3,4,5,6,7", "8:00", null);
                event = createEvent();
                event.patientDataEntry("4985","BG Fasting","numeric","1 hour");
                event.insert();
        }
}

unitProjection("20130","2 abnormal measurements in past week") {
        annotateTemporal("or",new String[] {
                        "event.getNumber(4985)>=150",
                        "event.getNumber(4986)>=150",
                        "event.getNumber(4987)>=150",
                        "event.getNumber(4988)>=150"
                }, "abnormal_BG", "date" );
        while (true) {
                waitTemporalQuery("count >= 2", "abnormal_BG", "8 calendardays");
                callback("5112", "2 abnormal values in BG were found in your
                        measurements in the past week,
                        system is calculating another schedule for you for daily BG measurement");
        }
}
```

**Figure 11**. An example of a projection file sent to the mobile decision-support system (mDSS) from the back-end decision-support system (BE-DSS), containing two unit-projections.

Before building a projection on the fly, the projection engine checks the current context of the patient (e.g., "Regular Schedule", which is induced by the "at work" event); it then retrieves all of the patient's scheduling preferences for this context (e.g., days and hours of reminders), and modifies the projections accordingly. For example, in unit-projection "20102", the time to activate reminders to the patient is set to "8:00" which is the preferred hour by the patient to get reminders in the context "Semi-routine Schedule".

In addition, the projection file contains two lists of IDs: one for the sub-plans to be stopped, and one for the sub-plans to be started. When the BE-DSS is triggered (e.g., by an incoming call-back from mDSS, or through detection of a breakout pattern), all affected sub-plans that are needed to transit into their complete state are aggregated by the projection engine into a unified *stop-list*. On the other hand, sub-plans that need to start are aggregated into a *start-list*. Thus, the mDSS is always "up-to-date" with respect to the sub-plans to be stopped or to be started at the local level (the rest of the plans currently applied by the mDSS are assumed by default to be continuing).

If the patient's mobile device crashes, the projection engine recovers the last procedural projections sent to the mobile device, and resends them to the mDSS. To support this functionality, we added to the BE-DSS a new *projected-plans* collection to store the different projections generated during the GL application session by the projection engine. Table 3 shows this collection in the context of the projection shown in Figure 11. Each row represents a unit-projection in the collection that has a link to the projection ID it belongs to, a unit-projection ID, the timestamp showing when it was sent to the mDSS, and status (started or stopped). Note that all unit-projections shown in Table 3 are sent from the same projection at the same time. The mDSS uses these properties to manage the execution of the sub-plans running locally (for example, it might stop the "daily blood pressure monitoring" sub-plan #22 from the stop-list, and start a new plan for "twice a week" monitoring sub-plan #23 from the start-list). The time sequence of these projections is shown in the sequence diagram is displayed in Appendix A.

**Table 3.** The *projected-plans* data structure, which stores the different unit-projections generated by the projection engine.

| Projection ID | Unit-projection ID | SentDate | Status |
| --- | --- | --- | --- |
| 184 | 20091 | 10/5/14/14:00:00:00 | stop |
| 184 | 20092 | 10/5/14/14:00:00:00 | stop |
| 184 | 20102 | 10/5/14/14:00:00:00 | start |
| 184 | 20130 | 10/5/14/14:00:00:00 | start |

### 2.3.3 Implementation of personalization through dynamic projections

At projection time, except for patient preferences for days and hours of reminders, the projection engine replaces all pre-defined knowledge thresholds with real values. An example of a knowledge threshold is the personal target level for physical exercise. Values above this threshold are abnormal. As the threshold values might be changed from time to time, writing their explicit value in the projection file will be hard to maintain. Instead, the threshold knowledge ID is written as a variable name (a string) circumscribed by triangular brackets; at projection time, the variable is replaced by the real value from the knowledge base. Figure 12 shows an example for unit-projection "20010": at design-time, the threshold knowledge ID is set, for example, to the variable "<$5066$>" (which in this case denotes the exercise target level). At projection time, the projection engine replaces this string with the real value; in this case, all thresholds are set to "5".

```
unitProjection("20010","Weekly METS") {
    while (true) {
        waitPeriodic("7", "7:00", null);
        if (temporalQuery("sum >= <$5066$>", "5065", "7 days")) {
            patientNotification("5162","Enhorabuena, el ejercicio
                                       ayuda al buen control. Sigue así.");
        } else {
            patientNotification("5163","Recuerda que hacer ejercicio
                                       es importante para tu bienestar
                                       y para mantener un buen control
                                       de la glucosa.");
        }
    }
}
```

**Figure 12.** The knowledge thresholds in the case of calculating the threshold for weekly exercise. Patient notification texts are shown in Spanish (for the gestational diabetes [GDM] pilot in Spain).

Another dynamic projection behavior is handling medication prescriptions: as medication prescriptions are patient-specific they cannot be part of the GL knowledge. Thus, the projection engine adds each valid medication (plus its appropriate dose and schedule) it finds in the patient's personal record dynamically as a unit-projection, and converts the dates to start and stop the medication to total days. Each medication is then personalized according to the current context of the patient; for example, as shown in Figure 13, in the case that the time for taking a medication is set to "after dinner", this time is generated according to the lunch hour belonging to the current context so that the mDSS receives the specific hour for taking medication, in this case 20:00. In addition, when the projection engine adds a medication to the stop-list it is not valid anymore.

```
unitProjection("60a978c3-d296-4d66-b0af-4bffe78fd220","Medication Take") {
        var dosages = {"20:00":"80.0 mg"};
        var reminders = {"20:00":"30.0 minutes"};

    while (true) {
            for (var time in dosages) {
                var dosage = dosages[time];
                var reminder = reminders[time];
                waitPeriodic("4,5,6,7,1,2,3",time,reminder,"0","61");
                setProjectionGlobal("AFDoseId", createUUID());
                event = createEvent();
                event.patientDataEntry("9648","Prendi il farmaco atorvastatina, " + dosage + " ","boolean","2 hours");
                event.insert();
                    ..........
        }
```

**Figure 13**. Projection of a medication therapy with personalization of the reminder time.

### 2.3.4 Preserving Robustness: Design and Implementation of the Recovery Mechanism

The distributed DSS architecture we designed for the MobiGuide project is very ambitious: On one hand, the BE-DSS should handle many requests concurrently, apply any relevant guideline personalizations in a dynamic fashion, generate projections, and send them back to the mDSS. On the other hand, the mDSS must be able to accept the projections at any time, and continuously interpret and apply them (e.g., on a daily basis).

Such an architecture, however, might suffer from unexpected failures: errors such as software bugs, software crashes at the mobile or at the server, or server overloadings might occur and cause failure of one of the components (or both). When this kind of error occurs, the BE-DSS should have the ability to recover and resume the GL application session from the last point of failure. This includes in our case also re-sending the relevant projections. For example, when the mDSS crashes, it might be necessary to re-install the application software in the mobile, and re-enroll the patient. In that case, the BE-DSS should re-send all of the most recently active projections, personalized to the specific patient, and continue the GL application session. Thus, we designed and implemented our architecture to be robust with respect to two main issues:

1. Ability to resume a session from the point at which it was stopped, including the sending of personalized projections after an error occurred in the patient's mobile device, such as a crash of the software.
2. Dynamic updating of the projections by considering the time elapsed since the last projections. For example, in Figure 13 the duration specified for taking the medication is two months, or 61 days (the default time unit used for projections in the AF domain). Assume that the patient's mobile device crashed after 30 days. Now, the BE-DSS should send the projection again, but with a specification of only a 31-days duration. Moreover, if the crash happened after more than 61 days, the projection should not be sent at all.

These two mechanisms ascertain that our architecture is recoverable and robust, regardless of the cause of the failure; our system will always recover, resume the application session, re-send the relevant projections, and continue to manage the patient.

As explained earlier, we have specified and applied the GDM and AF guidelines within the MobiGuide architecture, managing GDM patients in Barcelona, Spain, and AF patients in Pavia, Italy. Thus, in the next section we describe the precise evaluation methods of the current study, as these pertain to the two guidelines and to their continuous real time application to the two groups of patients.

# 3 The Evaluation methods

## 3.1 Evaluation of the specification process when using the PCB methodology

To assess the feasibility and validity of the new PCB model, we have specified, using domain experts and the new methodology, two guidelines that are very different, with respect to their population, intensity of monitoring, and application actions: GDM [54, 55] and AF [56].

The GDM patients were young, pregnant Spanish women, experienced with Smartphones and computers, who are otherwise healthy and have complications of pregnancy related to diabetes with or without hypertension. The AF patients were older, chronic patients, had additional comorbidities, and were much less experienced with computers and Smartphones. In addition, the AF and GDM domains vary in terms of intensity of monitoring: more intense monitoring of ECG in AF (1-2 daily sessions of 30 minutes each) versus up to four discrete measurements a day of the blood glucose (BG) level, in the case of GDM. Thus, the types of data collected, patterns monitored, the amount of interaction between the BE-DSS and mDSS, and the amount of different recommendations and notifications that it provided to patient and clinician users were quite different. The description of the data collection process in MobiGuide is out of the scope of the current paper and can be found in [39, 40].

For each GL, we defined several measures in the context of the knowledge specification language such as the number of each type of resultant knowledge roles in the knowledge base representing the specified guidelines. Table 4 describes the different measures for the evaluation and their description. We used the total # and its SD as the method to measure the first hypothesis.

**Table 4**: The different measures used in the evaluation study and their descriptions.

| Measure | Description | Example |
|---|---|---|
| Raw concepts | Patient data collected from various sources such as sensors and PHR | BG, Ketonuria, BP |
| (Temporal) data patterns | Abstracted pattern monitored contently by the Mediator | "two bad measurements of BG within the past week" |
| Conditions/Criteria | Eligibility or Abort retirees | Patient eligible for cardioversion therapy |
| Customized context | Different personal events inducing additional context on patient | "routine", "semi-routine" |
| Notifications to patients | Message to patient he doesn't need to accept or decline | "You wore the sensor for less than 6 hours. Please remember that the recommended monitoring time is 24 hours" |
| Notifications to care providers | Message to care-giver he doesn't need to accept or decline | "Patient is not compliant to 'pill in the pocket' procedure" |
| Recommendations to patients | Message to patient he needs to accept or decline | "Start HR sensor for 30 minutes" |
| Recommendations to care providers | Message to care-giver she needs to accept or decline | "Adjust insulin therapy for the patient" |
| Projections | Small protions of the GL downloaded to mobile | "Semi-routine daily BG management" |
| Callbacks from mDSS to BE-DSS | Data-notification from the mobile when BE-DSS needs to take control | "Call-back – abnormal BG in Daily schedule" |
| Monitoring plan projections | Plans to monitor mDSS Call-backs | Monitor Call-back abnormal BG in Daily schedule |

## 3.2 Technical and Pre-Clinical Evaluations of the PCB model

In order to make sure that the PCB model is operationally valid, i.e., that the system generates the correct recommendations at the correct time for the right actor (i.e., care provider or patient), we performed an extensive technical evaluation two months before the clinical pilot. The pre-pilot was performed using volunteers who agreed to test the system in its early phases, in both of the clinical domains: nine volunteers for the AF domain and seven for the GDM. This pre-pilot phase included a rigorous verification of the system's functionality on a set of simulated but highly realistic longitudinal patient records, such as checking for correct context switches, discovery of meaningful temporal patterns in the data, and making sure that they trigger the BE-DSS to change plans and project new projections to the mDSS. In cases of an inconsistent behavior (e.g., an incorrect timing of a recommendation, or a wrong notification to the patient), a team of four clinicians and three knowledge engineers worked on a daily basis to create a revised version of the computerized GL, in order to test the system's output again. Tests also included a verification that the hospital's EMR data is imported correctly to the MobiGuide PHR, an examination of all BAN-server activations (due to messages from the Picard engine, or due to data that are entered into the PHR and are sent to the mDSS through the BAN server), triggering of relevant monitoring parameters, and checking that all of the contents, and only the contents, of technical and clinical recommendations and reminders were correctly generated and sent to the appropriate patient and caregiver GUIs.

In order to bootstrap and scale the process of technical evaluation, we developed a simulation engine to test all of the different workflows covered by the GLs, and generated patient data for long periods of time. We used a technique for guideline-driven simulation of longitudinal patient records that was similar to the one that was originally used to rigorously test the Picard DSS engine in the Pre-eclampsia/Toxemia (PET) domain [11, 42]. A discussion of the simulation is out of the scope of the current paper. However, additional details can be found in Appendix B. The simulation of the data was mostly important in the case of the GDM GL, because the temporal patterns that were monitored spanned several weeks (e.g., 30 days of good blood glucose values).

## 3.3 Functional evaluation of PCB model implementation

The pilot itself involved ten AF patients from IRCCS Foundation Salvatore Maugeri, Pavia, Italy and twenty GDM patients from the Parc Tauli Sabadell University Hospital, Sabadell, Spain. The study began on April 1, 2015 and ended on December 31, 2015. One of the GDM patients (patient 11) dropped out after one week, so the results will be reported for the nineteen patients who participated in the full pilot.

To assess the PCB model during the pilot, we defined four different interaction types representing different functionality of the BE-DSS in the context of the pilot. We used the total# and SD as a method for each measure:

- **Data notifications** – a data notification occurs when the Temporal Mediator (the central temporal-reasoning engine) notifies the PICARD DSS (central GL engine) that a previous subscription to a particular pattern has just been triggered. Note that a Call-Back message from the mDSS, indicating that a certain temporal pattern was encountered by the mDSS, is a special type of a triggering pattern (a Call-Back pattern might also be computed and detected centrally by the Temporal Mediator). When the PICARD DSS is notified, several outcomes might result: sending projections, providing recommendations to the care-giver, or providing recommendations to the patient. In the context of the evaluation, we analyzed which events triggered and notified the BE-DSS, and in which clinical paths.
  Also, we defined and calculated for each patient an *"Interaction rate"* (normalized to days), or "*Functional Mean Time Between Interactions*" (*FMTBI*), which means, how many days, on average, passed between any two different guideline-related interactions with the BE-DSS. E.g., if a patient's mobile device had an MTBI of 4.8 on average, it means that every 4 to 5 days, on average, there was an interaction of the device's mDSS with the BE-DSS. We also defined "*Technical Mean Time Between Interactions*" (*TMTBI*), which refers to any kind of interaction with the BE-DSS, including interactions due to technical problems such as a crash of the mobile device.

- **Projections** – These are the unit projections sent to the mDSS by the BE-DSS. In the context of the evaluation we want to analyze how many and which projections are sent to the mobile during the pilot. Also, as each projection might potentially stop the current schedule of an on-going plan and start a new scheduled plan, we also want to analyze which plans were re-scheduled.
- **Care-giver recommendations** – these are recommendations or notifications to the care-giver. In the evaluation we want to analyze the compliance of the care-giver with those recommendations.
- **Patient recommendations** – these are recommendations, patient-data-entry, and notifications sent from the BE-DSS to the patient. In the context of the analysis we want to analyze the compliance of the patient with those recommendations.

In order to better analyze the results, for each interaction type (except for projection) we defined several sub-types. Table 5 describes the types, subtypes and their description.

**Table 5.** The types and sub-types of the analysis interactions in the MobiGuide System.

| Type | Sub-type | Description | Example |
|---|---|---|---|
| Data notifications | Callback triggered by the mDSS | The mDSS sends a message to the BE-DSS to "take control" and continue the application of the GL | The pattern of "two occurrences of High BG values in a week" |
| | Monitoring condition triggered | The Mediator notifies PICARD DSS about some pattern that triggered | The pattern of Monthly Good BG compliance |
| | Patient changed context | The patient changed his personal context in the mobile | Change context from "Routine" context to "Semi-routine" context |
| | Patient-data-entry entered | Survey data entered by the patient | Answering "no" to the question "There are some high glucose values, did you eat today more than you should have?" |
| | Care-giver accepted recommendation | Care-giver accepted recommendation | Respond "accept" the recommendation "Consider start of insulin treatment" |
| | Care-giver declined recommendation | Care-giver declined recommendation | Same as above with response "decline" |
| | Patient accepted recommendation | Patient accepted recommendation | Respond "accept" to the recommendation "Ketonuria has been positive; please increase your bedtime carbohydrates by 1 unit (10 grams)" |
| | Patient declined recommendation | Patient declined recommendation | Same as above with response "decline" |
| Care-giver recommendation | Procedure | Recommendation the care-giver should accept or decline | "Consider start of insulin treatment" |

| | Notification | Message the care-giver doesn't need to accept or decline | "Please make sure your patient's proteinuria is checked every month" |
|---|---|---|---|
| Patient recommendations | Procedure | Recommendation the patient should accept of decline | "Ketonuria has been positive; please increase your bedtime carbohydrates by 1 unit (10 grams)" |
| | Notification | Message to the patient he doesn't need to accept or decline | "Your last BG measurements weren't so good – you should visit your doctor in the next few days" |
| | Patient-data-entry | Survey data the patient should enter | "There are some high glucose values, did you eat today more than you should have?" [Answer yes/no] |

## 4 Results

### 4.1 Results of the specification process using the PCB model

Together with the expert physicians we specified the GDM [54, 55] and AF GLs [56] using the GESHER knowledge acquisition tool [46]. A complete discussion of the knowledge acquisition process is out of the scope of this paper and can be found here [39, 44]. Each GL required approximately three months to specify in detail, in collaboration with domain experts. Following that phase, we identified projected plans in the GL. Most of the projected plans were periodic and monitoring sub-plans. Projected periodic sub-plans are plans in which some action should be performed periodically by the mDSS.

Table 6 displays the distribution of the projected plans across both of the GLs, and their characteristics. In the GDM domain, we specified 22 periodic projections, 17 monitoring plan projections, and 16 callbacks. These numbers dramatically decreased in the case of AF GL: we specified 18 periodic plan projections, only two monitoring plan projections, and only two call-backs. We discuss the implications of these results in the Discussion, in Section 5.1.1.

**Table 6.** Difference in CIG characteristics between the two clinical domains

| | AF | GDM |
|---|---|---|
| Raw concepts | 100 | 300 |
| (Temporal) data patterns | 71 | 124 |
| Conditions/Criteria | 20 | 69 |
| Customized context | 4 (semi-routine, routine, 24h monitoring, increased physical activity) | 2 (semi routine, routine) |
| Notifications to patients | 7 | 10 |
| Notifications to care providers | 20 | 2 |
| Recommendations to patients | 5 | 1 |
| Recommendations to care providers | 18 | 7 |
| Periodic Plan projections | 18 | 22 |
| Monitoring plan projections | 2 | 17 |
| Call-backs from mDSS to BE-DSS | 2 | 16 |

## 4.2 Results of the technical evaluation

Although the AF guideline knowledge base was already ready before the pre-pilot,, the pre-pilot helped us to finalize the knowledge, and fix syntactic and semantic bugs in the guideline during the testing of the GL. The changes in GL knowledge included:

- Adding a medication non-compliance questionnaire to the GL
- Adding different recommendations for "pharmacological treatment" in the GL's caregiver part
- Changing translations and *Virtual Medical Record* (VMR) classes that should have been used to represent each data-item. These VMR classes were predefined and used to access the patients' data in the PHR. Those that were found to be incorrect were modified..
- Modifying the declarative projection so as to include support for recovery from a mobile crash.
- Modifying the procedural projections to support the valid time of each recommendation (medication or measurements).

Most of the changes made in the BE-DSS in the pre-pilot period were related to changes to the procedural projection and personalization mechanism, especially the aspects related to handling mobile crashes and correctly resuming the GL application. These changes were continuously tested , and if there were still bugs, they were reported.

To better facilitate this process, we developed and deployed during the pre-pilot phase a "projection log" viewer. For example,, we performed the following changes to the BE-DSS:

- Adding a new functionality for correctly restoring a medication projection as part of the capability to recover from mobile crashes
- Adding support for handling a context change by the user, and sending all active medications with a new personalized scheduling
- Support the capability to turn off the sending of reminders of any type

As a result of performing all of the necessary modifications, based on the comments of the volunteers, the clinicians, and the knowledge engineers, both guidelines operated, at least as used by the volunteers, in a satisfactory manner.

## 4.3 Results of applying the PCB model in the case of the GDM Guideline

In this section, we present the analysis of the functionality of GDM management in the context of the analysis of the BE-DSS functionality.

### 4.3.1 General analysis and interaction rate

Table 7 describes the duration of the overall interaction period for each patient using the DSS, and the number of interactions with the BE-DSS, which means that "data-notification" has been sent to PICARD to resume the GL execution. The average duration of the GL application was $61.68 \pm 20.55$ days. The number of interactions with the BE-DSS for each patient is also shown in Table 7. The total average number of interactions for all patients was $17.53 \pm 9.59$. Note that patient 7 had a relatively high number of interactions due to the fact that she changed her context at a high frequently (see explanation later in this section).

The rightmost column of Table 7 describes the functional interaction rate of each patient's mDSS with the BE-DSS in the GDM domain, the FMTBI (see Section 3.3). The average was $3.95 \pm 1.95$, which means that on average, every 4 days there was an interaction of the mDSS with the BE-DSS as part of applying the GL to this patient. Except for one time, there were no technical problems in the mobile; thus there was no need to calculate the TMTBI (see Section 3.3).

**Table 7.** Duration of each guideline-based decision-support session and interactions with the back-end decision-support system (BE-DSS), in the case of the application of the Gestational Diabetes (GDM) guideline.

| Patient ID | Days in MobiGuide | Number of data-notifications to the BE-DSS | Functional Mean Time Between Interactions (FMTBI) |
|---|---|---|---|
| 1 | 44 | 11 | 4.00 |
| 2 | 48 | 6 | 8.00 |
| 3 | 41 | 12 | 3.42 |
| 4 | 77 | 13 | 5.92 |
| 5 | 52 | 16 | 3.25 |
| 6 | 46 | 12 | 3.83 |
| 7 | 91 | 42 | 2.17 |
| 8 | 103 | 32 | 3.22 |
| 9 | 77 | 17 | 4.53 |
| 10 | 43 | 11 | 3.91 |
| 11* | -- | -- | -- |
| 12 | 55 | 12 | 4.58 |
| 13 | 59 | 11 | 5.36 |
| 14 | 76 | 20 | 3.80 |
| 15 | 22 | 7 | 3.14 |
| 16 | 92 | 34 | 2.71 |
| 17 | 55 | 13 | 4.23 |
| 18 | 55 | 22 | 2.50 |
| 19 | 72 | 24 | 3.00 |
| 20 | 64 | 18 | 3.56 |
| **Mean** | **61.68± 20.55** | **17.53± 9.59** | **3.95±1.35** |

*Dropped after one week.

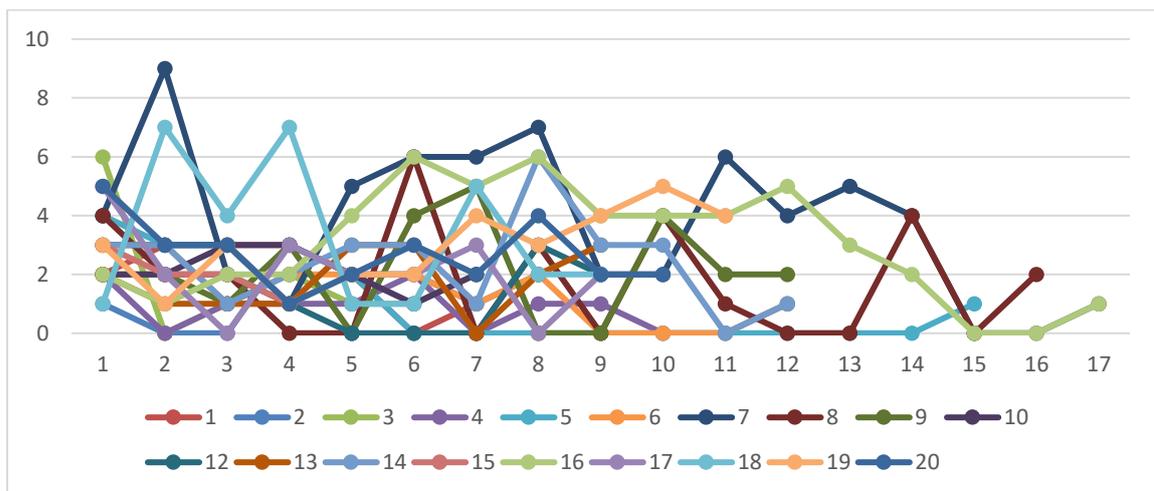

**Figure 1**4. Number of interactions per week of the local mDSS with the central BE-DSS over time (in weeks) for each patient in the Gestational Diabetes (GDM) domain.

Figure 14 displays the timeline of each patient, and the number of interactions with the BE-DSS. It can be seen that the mDSS interacted with the BE-DSS in each week, and mainly between weeks 5 and 9. This can be explained by the fact the patient took some time to get used to the application, then used it for another period of time, and then stopped using it, probably due to delivery. Each data-notification to the PICARD DSS resumes the GL application and therefore might potentially generate projections, patient recommendations, and care-giver recommendations.

Figure 15 presents the number of interactions for each patient in total. In general, most of the interactions with the BE-DSS were regarding the different data notifications. Several patients as well as their care-givers received recommendations. The distribution of the interaction types per each patient were the same: each patient had several interaction types of interactions.

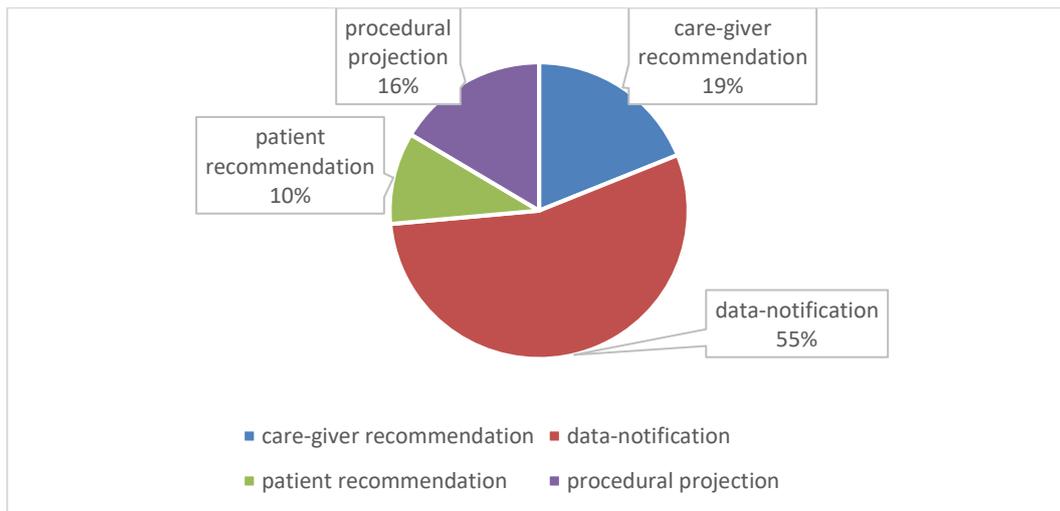

**Figure 15**: Distribution of interaction types across all patients in the Gestational Diabetes domain.

### 4.3.2 Data notification

With clinical context, most of the data-notifications to the BE-DSS were related to Blood Glucose (BG) management (87%), as expected considering that BG monitoring was the preferred functionality by patients and they were prescribed to monitor BG four times a day. The interaction with ketonuria was mostly performed at the mobile side by the mDSS, except when a diet therapy change was needed because of finding repeated patterns in positive ketonuria values (6%). Data notifications regarding BP were 4%, and data-notifications triggered to complete the GL due to baby delivery were in 3 of the cases.

Most of the interaction types to the BE-DSS were related to call-backs (44%) from the mDSS. All of the callbacks, except two, were related to the triggering of the pattern "2 abnormal BG in a week". Monitoring conditions also happened relatively frequently (22%). 18% of data notifications were due to care-giver declining recommendations, 8% were related to context change, 5% to patient-data-entry, and 2% were related to care-giver accepted recommendation. We discuss the implication of this result to distributed computing in Section 5.1.2.

Regarding the distribution of the monitored conditions that were actually triggered at the back-end during the GDM GL application: 39% of them were related to ketonuria monitoring, 30% were related to blood pressure monitoring, and 31% indicated monthly good compliance to blood-glucose measurement. The differences are probably related to the fact that ketonuria was measured on a daily basis, and the duration of this pattern was two weeks. The blood pressure was measured only twice a week, and the blood glucose measurement compliance pattern required a month of good compliance, although it was measured four times a day, so it was less likely to be triggered frequently.

## 4.4 Results of applying the PCB model in the case of the AF Guideline

This section describes the analysis results of the patient part of the pilot in the context of BE-DSS. In general, the AF guideline was designed to mostly be run by the mDSS, thus most of the interactions were related to projections or context change rather to call-backs or recommendations to the patient (call-backs were only modeled in BP monitoring plans). The reason for having mainly projection interactions is because each time the care-giver updated patient data (e.g., prescribe new medications), the guideline was re-started, and new projections are sent to the mobile. Also, when a new version was updated on the mobile, or when some technical problems were detected, projections were sent again. Data-notification such as context change also generated new projections to the patient. Thus, in the context of the evaluation we want to check how many and which types of projections were sent to patient, and the MTBI rate of each patient.

### 4.4.1 General analysis and interaction rate

Table 8 presents the general duration and the MTBI rate for each patient. Note that the table includes two MTBI scores: A Functional MTBI that includes only real data-notifications to the BE-DSS, and a Technical MTBI that also includes additional data notifications to the BE-DSS to re-send a projection when the mobile device crashed. It can be seen that overall, the patients spent a significant amount of time using MG (a mean of 127 days), with a relatively low mean Functional MTBI of 23.8 days. In other words, an average of more than three weeks passed between one call-back to another, between which the patients were managed through solely the mDSS. This interesting result is discussed in Section 5.1.3.

Figure 16 shows the interaction timeline for each patient. Again, like the GDM timelines, it can be seen that the first few weeks are hectic, with a lot of interaction, but after week 14 interactions are more or less stable, with only few interactions. Note that patient 7 had a peak in week 15; however this was caused by technical problem.

**Table 8.** The duration, FMTBI, and TMTBI for each patient when applying the AF guideline.

| ID | days | Functional data Notification | Functional MTBI | Technical data Notification | Technical MTBI |
|---|---|---|---|---|---|
| 1 | 96 | 6 | 16.00 | 7 | 13.71 |
| 2 | 78 | 5 | 15.60 | 6 | 13 |
| 3 | 98 | 3 | 32.67 | 6 | 16.33 |
| 4 | 91 | 4 | 22.75 | 9 | 10.11 |
| 5 | 90 | 2 | 45.00 | 3 | 30 |
| 6 | 136 | 6 | 22.67 | 7 | 19.42 |
| 7 | 259 | 23 | 11.26 | 43 | 6.02 |
| 8 | 89 | 9 | 9.89 | 11 | 8.09 |
| 9 | 249 | 13 | 19.15 | 26 | 9.57 |
| 10 | 86 | 2 | 43.00 | 4 | 21.5 |
| Mean±SD | 127.20±68.62 | 7.30±6.46 | 23.80±12.47 | 12.2±12.62 | 14.78±7.27 |

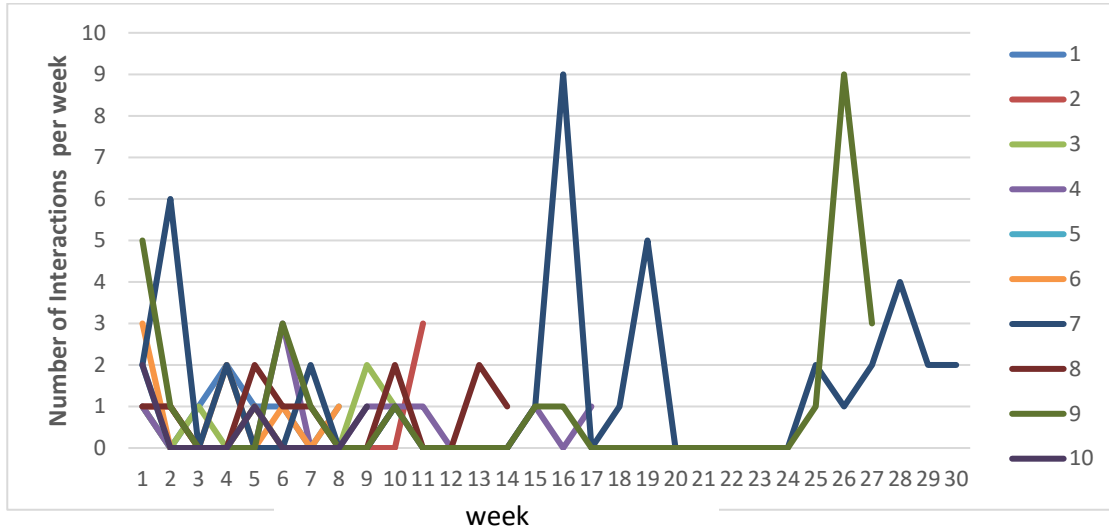

**Figure 16.** Number of interactions per day across the timeline of all patients in the atrial fibrillation domain.

### 4.4.2 Interaction types

Most of the interaction types were due to projections to the mDSS (83%); others were data notifications, mostly to change context (17%). Some of the data notifications were triggered due to technical errors.

For all patients, most of the projections originated due to the initial starting of the GL, but due to a technical problem in the mobile devices that led to frequent crashes (53%), initialization projections were often re-sent to the mobile after recovery (30%), demonstrating the importance of robustness in the architecture. 13% of interaction types were related to change in context by the patient, and 4% to data notification.

## 5 Summary, Discussion, and Main contributions

### 5.1 The PCB methodology versus current approaches

Traditional medical decision-support architectures provide decision support (to both patients and clinicians) only through centralized DSSs. Other, more recent solutions, follow the other extreme option, and focus only on mobile device applications. Neither solution, however, easily caters to the combination of multiple local monitoring actions that require significant computational resources, to local reminders, to the need to access central resources that cannot be delegated to a mobile device, such as an EMR or a (possibly changing) clinical GL library, or to the ensuring of the robustness of the system with respect to connectivity.

Thus, in this study we designed, implemented, and evaluated across several EU countries, a novel methodology for distribution of a medical DSS. The main objectives of this research were to show the feasibility of a GL specification process using the new *Projection-CallBack* (PCB) methodology, to implement the PCB model in a real GL-based DSS environment, and to demonstrate its actual clinical feasibility, at least in the case of the GDM and AF guidelines. To fulfil these objectives, we performed the following tasks: 1) Definition of the characteristics of distributed clinical DSSs; 2) Review and comparison of the relevant distributed DSS frameworks; 3) Design of a dynamic distributed DSS Methodology; 4) Design and implementation of an architecture for distributed DSS framework; 5) Definition of an evaluation methodology for distributed DSS applications; 6) Implementation of the PCB methodology; 7) Evaluation of the results through a pilot of specification and actual application of two GLs, in two clinical domains, each in a different country.

Unlike standard distributed-processing methods, *the algorithmic, knowledge-based contents of each patient's personal management unit change over time*. In addition, as we have explained in detail, the distribution of responsibilities between the central and local (patient) units follows certain well-defined guidelines and policies that we had outlined clearly. In the approach we had describe in this study, we focused mainly on the *personalization of the computation* performed by the personal device of each patient, based on their longitudinal, multivariate personal data (e.g., to determine whether a Call-Back pattern was detected), and on the specific evidence-based knowledge dynamically projected to that device from the central server (dependent on the patient's data), and *not* just on increasing the efficiency of the computation by using multiple processors, as is often the case for distributed-computation frameworks.

Note that the temporal data-based Call-Back patterns that govern *when* each personal patient-associated mobile device calls the central computer for further assistance are customized to each guideline and patient-specific context; and so are the context-sensitive guideline segments projected in response to the Call-Back request, from the central server to each specific patient-associated mobile device. It might be worth emphasizing again that such a conceptual framework is quite different from splitting a large computational process into smaller segments and running it on multiple separate, but essentially identical, local clients.

We now briefly summarize the results of the PCB model evaluation in the next subsections.

### 5.1.1 The specification process using the PCB model

As was shown in Table 6, the number of plan projections in the two domains (all of which happened to be periodic plans) was quite large (18 for the AF GL and 22 in the GDM domain), and the number of monitoring projections was not very small either (2 or 17, respectively). Thus, we have shown that multiple actions of the GL, especially measurements (such as BG measurements), can indeed be specified using the PCB model and projected to the mDSS, significantly reducing the computational load on the BE–DSS.

However, we also noted after the specification was completed that there is a significant difference between the characteristics of these two domains in terms of the Call-Back mechanism: in the case of the GDM GL, there are 16 call-backs and 17 monitoring plans, and in the case of the AF GL, there are only two call-backs and two monitoring plans. This means that most of AF GL's actions can be handled by the mobile device, which thus very rarely needs the BE-DSS to change the projection. In contrast, in the case of the GDM GL, more decisions are made at the BE-DSS level, since more decisions in that GL require additional data (such as past and future visits to the clinic) and care-giver confirmations, both of which are accessible only to the BE-DSS. This difference between the two GLs indeed manifested itself in an average FMTBI of $3.95\pm1.95$ days in the case of the GDM domain, versus a mean FMTBI of $23.80\pm12.47$ days in the case of the AF domain. Of course, a GL might have a small number of call-backs, which, however, occur frequently, and vice versa; but it seems that the number of call-backs might provide a reasonable estimate of the expected number of interactions between the local and central DSSs. In other words, it seems that a knowledge engineer can obtain a pretty good estimate of the characteristics of the GL when it is distributed, starting with the very first phase of GL elicitation and specification. For example, in our case, in terms of a distributed DSS it can be seen even in each of the GL knowledge bases' statistics after using the PCB model, that the AF GL is mostly controlled locally by the mDSS, whereas the GDM GL is mostly controlled centrally by the BE-DSS. This insight can be obtained even before applying the GL and measuring the distribution of projections and call-backs in a more quantitative fashion, such as through the FMTBI.

### 5.1.2    Applying the PCB model in the case of the GDM Guideline

As mentioned, we have found that patients in the GDM domain had an average FMTBI of 3.95±1.95 days (see Table 7). This period between successive local device and BE-DSS interactions (including in each interaction a call-back and a respective projection) implies that during that period, the mDSS has been managing the patient, saving the BE-DSS four days of computation. Table 7 also shows that over a period of 61 days, there were on average 17.53±9.59 interactions between the mDSS and the BE-DSS; thus, the computation was indeed distributed between the mDSS and the BE-DSS, especially in the case of data measurements such as BG, ketonuria, and BP, which are some of the basic management plans in the GDM GL. Delegating main functionalities of the GL such as planning for daily measurements to the mDSS can save a significant amount of computation on the BE-DSS, which can be exploited to handle less frequent situations, such as a context switch or sending notifications to the care-giver when an abnormal pattern is detected. Thus, in the case of the GDM domain, we have demonstrated the feasibility of using the PCB model to create a meaningful computational distribution between the mDSS and the BE-DSS.

### 5.1.3    Applying the PBC model in the case of the AF Guideline

In the case of the AF GL, we showed (see Table 8) that the FMTBI was 23.80±12.47 days. Thus, only every 23 days on average was there a need for an interaction of the mDSS with the BE-DSS; during these extended periods, the mDSS handled most of the GL-based management locally. This result has a significant implication in terms of computational resource saving. This behavior, although not necessarily directly implied, was suspected already during the GL elicitation and specification phase, due to the need to define only a small number of call-backs in the GL, as explained in section 5.1.1. This "profile" of the AF GL might be explained by examining the AF GL's clinical semantics: most of the periodic plans projections were related to medication administrations, ECG monitoring, and other plans that can be managed locally by the mDSS without the need to make a decision in the central sever.

In addition, the robustness and fast recovery of the runtime application, supported by the PCB model, were demonstrated here clearly, due to the relatively high number of mobile-device crash events. Thus, we have shown that applying the PCB model in the AF domain has also demonstrated the feasibility of creating a robust real-time distributed architecture.

### 5.2 The Main Contributions of the Current Study

The current study has provided three main contributions for the key Artificial Intelligence in Medicine task of automated, distributed chronic-patient management:

- o   Enhancement of the *knowledge modelling* and *knowledge-acquisition* process;
- o   *Design and implementation* of a new architecture for distributed guideline-based DSSs;
- o   Demonstration of the *real-world feasibility* of a fully distributed GL-based DSS.

The following subsections describe each of these contributions.

### 5.2.1 Enhancing the knowledge modelling and acquisition process to support distributed application of the GL

GLs are typically intended to be used by physicians. Patients at home are either completely left out of the decision-support process, or use mobile-device applications that do not represent a continuous application of an evidence-based guideline. Thus, enhancing the "traditional" GL specification process, with additional knowledge roles (such as the "*is projected*" property), and particularly adding in the process of determining which tasks might be best delegated to the local device, might bridge this gap. It might also provide a more realistic representation of the GL, which is more suitable for chronic patients who are mostly at home. In addition, after specifying the GL using the PCB model, the knowledge engineer might be provided with useful insights regarding the profile of the distributed computation of the GL. Thus, we demonstrated that in the case of GDM domain, most of the computation was performed on the BE-DSS, whereas in the AF domain, most of computation was performed in the mDSS.

The projection and call-back approach is a novel one also for GL specification, and provides new ways to characterize clinical GLs in a modern era, in which both centralized servers residing in the Cloud and local mDSSs residing on smart phones need to interact.

### 5.2.2 The design and implantation of a new architecture for distributed guideline-based DSSs

The design of the PCB model and particularly its underlying projection language supports a continuous dialog between the BE-DSS and the mDSS, and exploits the relative advantages of the different computational architectures and their respective access to clinical data and medical knowledge. This design, in which the central server can take over when needed and restart the process through projection, also supports in a robust fashion an almost continuous level of care, even when the patient is disconnected due to failure of the local device. Also, using measures to quantify the nature of the distributed process, such as the MTBI, might be useful to benchmark frameworks that support distributed DSSs.

### 5.2.3 Demonstration of the feasibility of a fully distributed GL-based DSS

We implemented the PCB model in two different domains. This implementation has demonstrated the feasibility of distributing the computational process of applying a complex GL between the mDSS and the BE-DSS. The demonstration included the feasibility of projecting and delegating portions of the GL to the mDSS, of monitoring for call-backs at the BE-DSS, and of changing the projection once a call-back is triggered. The feasibility of this model to support a distributed mode of GL-based decision support, and the fact that mobile devices today have very high computational power, enables in the future further implementation of different complex projection strategies.

### 5.3 Additional implications of the study

The current study was focused mainly on the innovative technical aspects of the PCB model; however, there are indications that architectures such as the MobiGuide project, which was powered by the new PCB technology, might potentially lead to several additional outcomes from the point of view of the patients and the care-givers, although we did not directly assess them in the current project:

- Contribution to patients – Patients using a distributed architecture are empowered by a greater awareness of the medical processes through which they are being managed, are more involved in their own management, and need to take more responsibility for their own care. For example, a distributed DSS can provide patients with notifications and alerts when needed, and this involves them more in the process, leading to a potentially higher level of compliance. Indeed, a preliminary indication of the high compliance of the patients in the MobiGuide pilot has already been pointed out [39, 40]. Furthermore, the application of the GL to each patient can be more easily customized and personalized using the distributed model, as demonstrated in another MobiGuide study [39, 40].

- Contribution to physicians – A distributed DSS framework can be used as a training and educational resource for clinicians who provide assistance to chronic patients in settings that involve the patient being at home or in other "real-life" contexts. This training can be provided by performing simulated applications of different GLs within multiple simulated out-of-hospital settings. Thus, just as a pilot trains herself on a flying simulator, a physician managing chronic patients in the community can use this distributed framework as a GL simulator. Such an educational resource might increase the professionalism of training physicians and increase their readiness to treat real cases.

## 6      Study Limitations and Future Directions

We have implemented and evaluated a comprehensive distributed DSS architecture; however, there is still a need to further investigate and support the following challenges in more depth:

- Number of patients and time of the experiment – because of the limited resources, the MobiGuide pilot included only 20 patients in the case of the GDM domain, and 10 patients in the case of AF (albeit for relatively extended periods). Further research using similar distributed architectures might include a larger number of patients, and perhaps even increase the duration of the follow-up period (e.g., a year instead of several months). It might thus be able to also assess the actual clinical outcomes of distributed management. The initial clinical results in the case of the MobiGuide project were quite encouraging. However, they were not statistically meaningful, although they certainly demonstrated a high patient compliance [40].

- Applying multiple GLs and handling comorbidities – Although in the current evaluation we did specify and apply two complex GLs, one of them (GDM) including also a potential comorbidity (hypertension), a future study should attempt to include GLs that handle several comorbidities. This would add significant complexity to the application, because each comorbidity is potentially handled by a GL that would need to be applied in parallel to the patient, sometimes with contradictions or redundancies between the GLs.

- The process of determining which tasks might be best delegated to the local device is currently performed manually, using the characteristics listed in Table 1. In the future, we might develop an automated process to suggest projection points within the GL during the specification phase, or following it.

- Support of the distributed GL application by a cluster of servers – It is considered desirable these days that every computational component should operate on a cluster of servers in distributed fashion; thus, in the future, this framework should operate as a set of components running on a cluster of servers. There are several options for optimal distribution of a GL application process, as demonstrated even when examining in depth only the continuous monitoring and interpretation task of multiple patients over time [58].

## 7      Conclusions

We have demonstrated in this study the feasibility of the PCB distributed DSS model, from specification to a full-fledged real-life clinical application. This feasibility has major implications for remote management of chronic patients: it allows central DSSs to delegate much of the monitoring process and treatment management decisions to the local mobile device, keeping the patients in their usual environment. Thus, in the future we anticipate the area of clinical DSSs to transform from the use of traditional centralized DSS architectures, to the use of distributed DSS architectures that can provide robust decision support to both the patients and their care providers, anytime, anywhere. We have provided two encouraging examples of the new distributed GL-application approach through the MobiGuide project.

**Acknowledgements.** The MobiGuide project (http://www.mobiguide-project.eu/) has received funding from the EU's Seventh Framework Programme for Research, Technological Development and Demonstration under grant agreement no. 287811. Partial support to Dr. Shalom was provided through the Josef Erteschik Chair of Information Systems Engineering at Ben-Gurion University.

# Appendix A: A time sequence diagram of the projection process

Figure A1 shows the time sequence diagram of the projection process: After patient enrollment, the first declarative projection file is sent to the mDSS using the pre-defined API. The resulting success message is sent to the BE-DSS to acknowledge the transition, and to start the GL application. The GL application might then trigger a second procedural projection, which is transmitted to the mDSS. During the GL application, several additional projection files might be sent to the mDSS according to the clinical context of the patient. Note that if there is failure or timeout, the BE-DSS tries to send the file again.

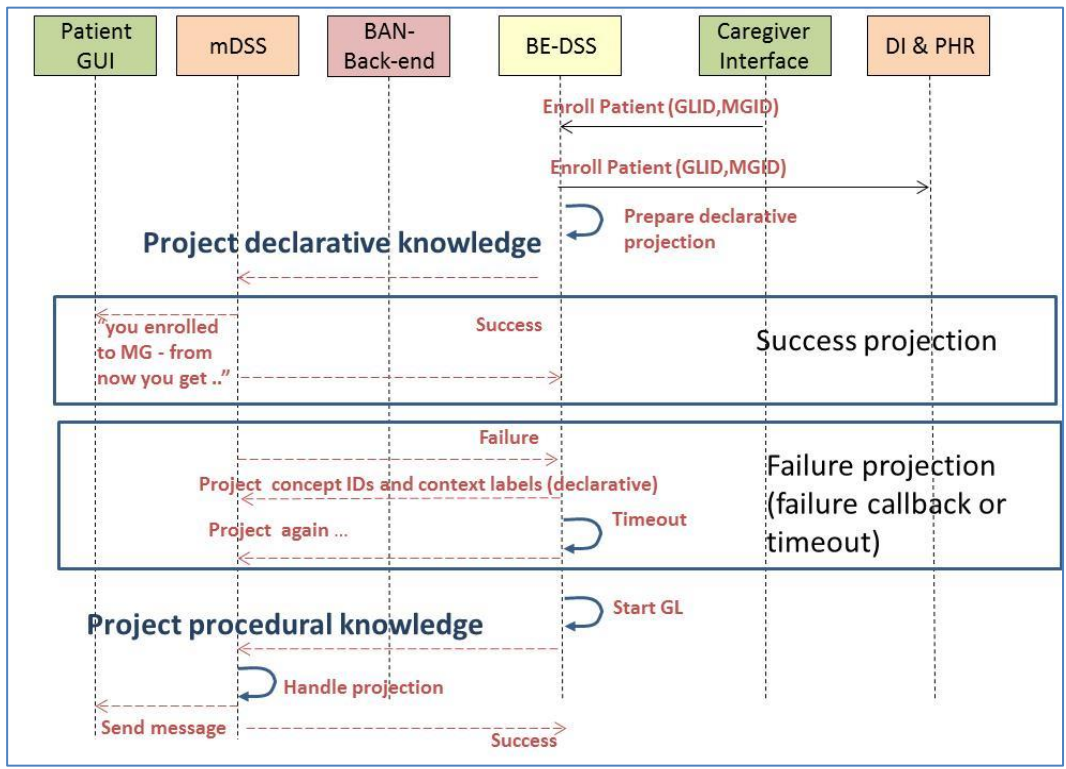

**Figure A1.** Time sequence of the projection process.

**Appendix B: Simulation of the GL application**

To test and validate the application and integration among all system components of the MobiGuide architecture as close as possible to the real environment, we decided to perform a comprehensive simulation of the GL application. This simulation includes the following tasks:

1. Defining the simulation scenarios
2. Defining the simulation data steps
3. Applying the simulation records

**1. Defining the simulation scenarios**

As a first step, in the case of the GDM GL, we specified in collaboration with a senior expert biomedical engineer a well-defined clinical scenario that occurs rather commonly when applying the GDM GL. We defined a virtual patient called "Molly" who uses the MobiGuide system. The scenario included all of the personal events of the patient, the demographic data, and seven weeks of treatment using the MG system – from week 27, when the patient enrolls to MobiGuide to week 34, when the patient delivers. Figure B1 shows the portion of the textual scenario created by the clinical expert and knowledge engineer.

As shown in Figure B1, each week is composed of daily activities the patient should perform using MobiGuide equipment. For example, during week 28 she performs regular blood glucose, ketonuria, and blood pressure monitoring. However, during the first day in week 29 she receives a compliance message from the MG system regarding her exercise, and after two days, the blood pressure monitoring scheduling is decreased to "once a week on Mondays", which is her preferred day. The change of the scheduling makes a new projection with a new personalized scheduling the mDSS received from the BE-DSS during the simulation, because the MobiGuide system detected she was compliant for a month. Thus during the specification of the scenario, we defined a lot of "MG moments" in which the MobiGuide system should react to changes in the data transactions. The next step was to define the dataset and its transactions.

**2. Defining the simulation data steps**

After defining the textual scenario, we simulated the longitudinal dataset. First, the knowledge engineer created a "gold-standard" dataset. A set of transactions, representing multiple time points within a single longitudinal patient record, was generated for each week in the GL scenario. Figure B2 shows a portion of the simulated dataset: for each transaction, the gestational week, day, time, the concept and its value are indicated. The dataset included all the transactions that should be inserted from all MobiGuide components, divided by steps. Each step represents a different MobiGuide component that should insert the data. The right column indicates the MobiGuide source that should simulate this transaction. For example, for step 3 all data should be inserted to the PHR from the smartphone GUI. We defined approximately 330 transactions for the complete the scenario using Microsoft Excel. At the end of each step, the MobiGuide system should react and send some recommendation. For example, after step 5 is generated, the mDSS should notify the patient about the importance of diet, because it detected the pattern "2 non-compliance diet entries in a week" (step 5a in Figure B2). The data for this pattern was pre-defined in step 5 according to the GL timeline. Also, the dataset includes the Virtual Medical Record (VMR) class that should be used to represent each data-item. These VMR classes were predefined. In the next step, each transaction should be automatically generated by a dedicated simulation engine developed in the mDSS.

### Part3 - At home (week 28)

1. Recommended amount of MEDS calculated and patient is recognized accordingly.
2. Patient omits one measurement of BG during this week, nothing happens
3. Patient omits one measurement of Ketonuria, nothing happens

**Week28.1: Sunday**
- Measure of BG **3** times a day, ketonuria
- *Amount of MEDS is calculated* and since she exercised well, receives a msg of recognition

**Week28.2: Monday**
- Measure of BG 4 times a day, ketonuria and **BP**

**Week28.3: Tuesday**

**Week28.4: Wednesday**
- Measure of BG 4 times a day, ketonuria

**Week28.5: Thursday**
- Measure of BG 4 times a day, ketonuria and **BP**

**Week28.6: Friday**
- Measure of BG 4 times a day, ketonuria

**Week28.7: Saturday**
- Measure of BG 4 times a day, ketonuria

### Part4 - At home (week 29)

1. On Tuesday we close two weeks of good BP – she moves to once a week measuring
2. One bad BG measurement on Monday, nothing happens
3. She didn't exercised at all in past week, and receives a reinforcement message

**Week 29.1: Sunday**
- Measure of BG 4 times a day, ketonuria
- *Amount of MEDS is calculated* and since she DIDN'T exercised well, receives a reinforcement message

**Week29.2: Monday**
- Measure of BG 4 times a day (1 BAD lunch), ketonuria and **BP**
- She exercises

**Week29.3: Tuesday**
- Measure of BG 4 times a day, ketonuria
- *Two weeks of good BP* – monitor moves to once a week based on preferred day (MON)

**Week29.4: Wednesday**
- Measure of BG 4 times a day, ketonuria
- She exercises

**Week29.5: Thursday**
- Measure of BG 4 times a day, ketonuria
- *Ketonuria is negative for 2 weeks* -> move to twice a week monitoring (MON, Thursday)

**Week29.6: Friday**
- Measure of BG 4 times a day, she exercise

**Week29.7: Saturday**
- Measure of BG 4 times a day,

**Figure B1**. Portion of the textual scenario defined for guideline simulation in the case of the gestational diabetes (GDM) guideline.

| week | day in week | day of treatment | valid time | GESHER ID | VMR Class | conceptName | Valid Start Time | Valid End Time | value | Steps | GENERATED BY component |
|---|---|---|---|---|---|---|---|---|---|---|---|
| 27 | 2 | 2 | 20:00 | 4988 | 8 | BG_dinner | 2/3/2014 20:00:00 | 2/3/2014 20:00:00 | 70 | | |
| 27 | 3 | 3 | 8:00 | 4985 | 2 | BG_fasting | 3/3/2014 8:00:00 | 3/3/2014 8:00:00 | 85 | | |
| 27 | 3 | 3 | 8:00 | 5021 | 29 | Ketonuria | 3/3/2014 8:00:00 | 3/3/2014 8:00:00 | -- | | |
| 27 | 3 | 3 | 9:00 | 4986 | 4 | BG_breakfast | 3/3/2014 9:00:00 | 3/3/2014 9:00:00 | 160 | | |
| 27 | 3 | 3 | 13:00 | 4987 | 6 | BG_lunch | 3/3/2014 13:00:00 | 3/3/2014 13:00:00 | 105 | | |
| 27 | 3 | 3 | 16:00 | 5065 | 95 | MET | 3/3/2014 16:00:00 | 3/3/2014 16:00:00 | 5 | | |
| 27 | 3 | 3 | 20:00 | 4988 | 8 | BG_dinner | 3/3/2014 20:00:00 | 3/3/2014 20:00:00 | 108 | 3 | SmartphoneGUI |
| 27 | 4 | 4 | 8:00 | 4985 | 2 | BG_fasting | 4/3/2014 8:00:00 | 4/3/2014 8:00:00 | 85 | | |
| 27 | 4 | 4 | 8:00 | 5021 | 29 | Ketonuria | 4/3/2014 8:00:00 | 4/3/2014 8:00:00 | ++ | | |
| 27 | 4 | 4 | 9:00 | 4986 | 4 | BG_breakfast | 4/3/2014 9:00:00 | 4/3/2014 9:00:00 | 160 | | |
| 27 | 4 | 4 | 13:00 | 4987 | 6 | BG_lunch | 4/3/2014 13:00:00 | 4/3/2014 13:00:00 | 105 | | |
| 27 | 4 | 4 | 20:00 | 4988 | 8 | BG_dinner | 4/3/2014 20:00:00 | 4/3/2014 20:00:00 | 108 | | |
| 27 | 5 | 5 | 8:00 | 4985 | 2 | BG_fasting | 5/3/2014 8:00:00 | 5/3/2014 8:00:00 | 85 | | |
| 27 | 5 | 5 | 8:00 | 5021 | 29 | Ketonuria | 5/3/2014 8:00:00 | 5/3/2014 8:00:00 | ++ | | |
| 27 | 5 | 5 | 8:00 | 5177 | 96 | SBP | 5/3/2014 8:00:00 | 5/3/2014 8:00:00 | 120 | | |
| 27 | 5 | 5 | 8:00 | 5178 | 96 | DBP | 5/3/2014 8:00:00 | 5/3/2014 8:00:00 | 70 | | |
| 27 | 5 | 5 | 8:03 | 5037 | 39a | patient data entry : Are you Eating enough carbohydrates ? | 5/3/2014 8:03:00 | 5/3/2014 8:03:00 | yes | 3a | mDSS (sending message to SmattPhoneGUI) |
| 27 | 5 | 5 | 8:04 | 5169 | 44 | Call back eating enought carbohydrates | 5/3/2014 8:04:00 | 5/3/2014 8:04:00 | TRUE | 3b | mDSS (saving in PHR callback) |
| 27 | 5 | 5 | 8:05 | 5051 | 41 | Increase carbohydrates at dinner/bedtime | 5/3/2014 8:05:00 | 5/3/2014 8:05:00 | TRUE | 4 | BE-DSS (sending recommendation to mDss) |
| 27 | 5 | 5 | 8:05 | 5051 | 42 | Increase carbohydrates at dinner/bedtime | 5/3/2014 8:05:00 | 5/3/2014 8:05:00 | TRUE | 4a | mdss/ Smarthphone (accepts recommendaiton) |
| 27 | 5 | 5 | | 4986 | 4 | BG_breakfast | 5/3/2014 9:00:00 | 5/3/2014 9:00:00 | 160 | | |
| 27 | 5 | 5 | 13:00 | 4987 | 6 | BG_lunch | 5/3/2014 13:00:00 | 5/3/2014 13:00:00 | 105 | | |
| 27 | 5 | 5 | 16:00 | 5065 | 95 | MET | 5/3/2014 16:00:00 | 5/3/2014 16:00:00 | 5 | | |
| 27 | 5 | 5 | 20:00 | 4988 | 8 | BG_dinner | 5/3/2014 20:00:00 | 5/3/2014 20:00:00 | 108 | | |
| 27 | 6 | 6 | 8:00 | 5021 | 29 | Ketonuria | 6/3/2014 8:00:00 | 6/3/2014 8:00:00 | -- | | |
| 27 | 6 | 6 | 8:00 | 4985 | 2 | BG_fasting | 6/3/2014 8:00:00 | 6/3/2014 8:00:00 | 85 | | |
| 27 | 6 | 6 | 9:00 | 4986 | 4 | BG_breakfast | 6/3/2014 9:00:00 | 6/3/2014 9:00:00 | 120 | | |
| 27 | 6 | 6 | 13:00 | 4987 | 6 | BG_lunch | 6/3/2014 13:00:00 | 6/3/2014 13:00:00 | 105 | | |
| 27 | 6 | 6 | 19:30 | 5055 | 49 | Nutrition management | 6/3/2014 19:30:00 | 6/3/2014 19:30:00 | ++ | 5 | SmartphoneGUI |
| 27 | 6 | 6 | 20:00 | 4988 | 8 | BG_dinner | 6/3/2014 8:00:00 | 6/3/2014 8:00:00 | 108 | | |
| 27 | 7 | 7 | 8:00 | 5021 | 29 | Ketonuria | 7/3/2014 8:00:00 | 7/3/2014 8:00:00 | -- | | |
| 27 | 7 | 7 | 8:00 | 4985 | 2 | BG_fasting | 7/3/2014 8:00:00 | 7/3/2014 8:00:00 | 85 | | |
| 27 | 7 | 7 | 9:00 | 4986 | 4 | BG_breakfast | 7/3/2014 9:00:00 | 7/3/2014 9:00:00 | 120 | | |
| 27 | 7 | 7 | 12:30 | 5055 | 49 | Nutrition management | 7/3/2014 12:30:00 | 7/3/2014 12:30:00 | ++ | | |
| 27 | 7 | 7 | 12:32 | 5171 | 65 | Notify patient about importance of following diet | 7/3/2014 12:32:00 | 7/3/2014 12:32:00 | TRUE | 5a | mDSS (sending message to SmattPhoneGUI) |

**Figure B2**. A portion of the gestational diabetes (GDM) guideline dataset transactions.

## 3. Applying the simulation records

The mDSS is automatically started by the Smartphone GUI, thus no manual intervention is required to operate the mDSS. However, to run a simulation, the user needs to select any simulation step to run from within the Smartphone GUI (Figure B3a). The simulator will then step through a series of events one by one, ignoring the exact step number selected (Figure B3b).

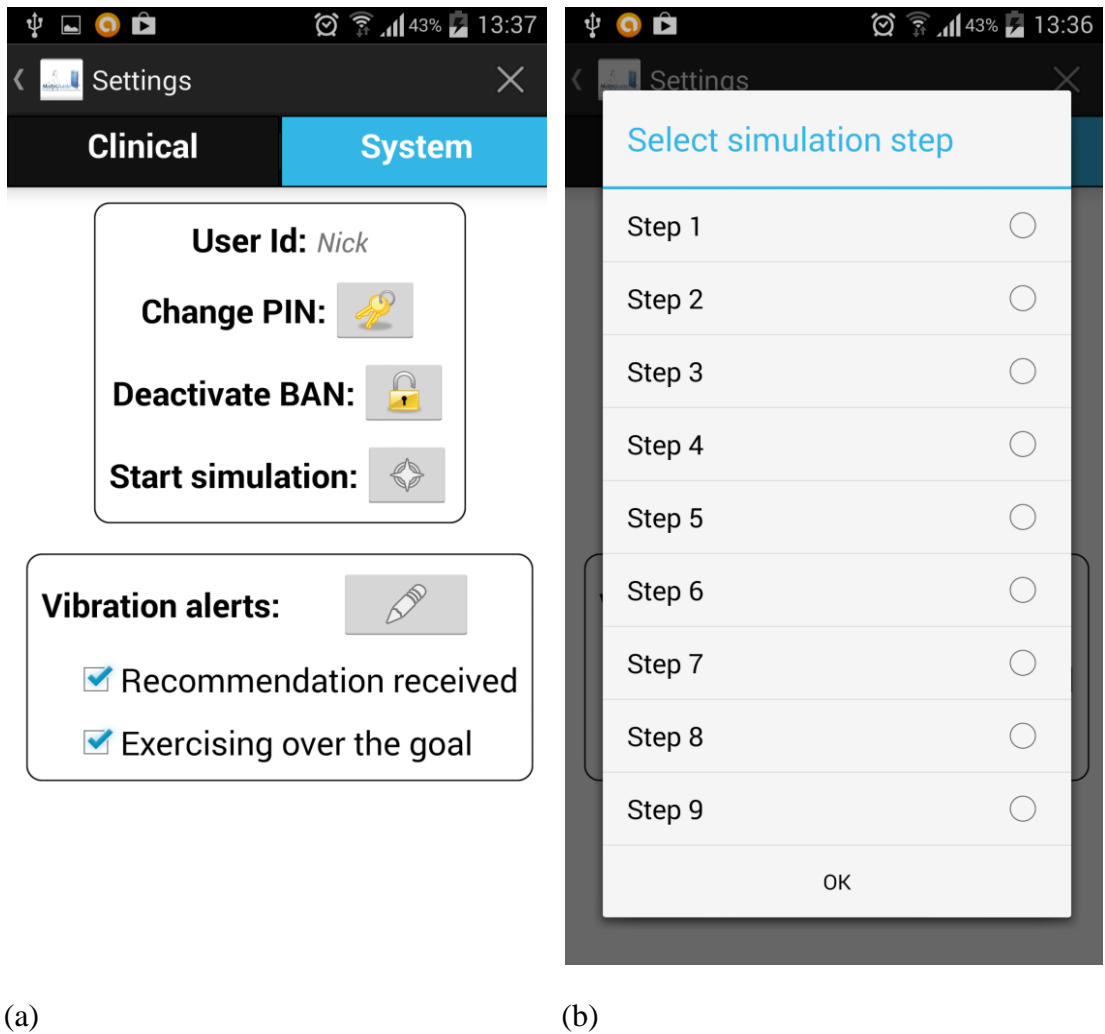

(a)     (b)

**Figure B3.** Screenshots of the Smartphone graphical user interface (GUI) showing: (a) the Settings page for accessing the simulation and (b) the Simulation Step page for running the simulation. Note that the specific step number selected has no effect on the simulation